%% file: main_arxiv.tex
\documentclass[10pt,twocolumn,letterpaper]{article}
\usepackage{authblk}

\usepackage[pagenumbers]{cvpr} %

\usepackage{graphicx}
\usepackage{amsmath}
\usepackage{amssymb}
\usepackage{mathtools}
\usepackage{booktabs}
\usepackage{nicefrac}
\usepackage{multirow}
\usepackage{subcaption}
\usepackage[table,xcdraw]{xcolor}
\usepackage[pagebackref,breaklinks,colorlinks]{hyperref}

\usepackage[capitalize]{cleveref}
\crefname{section}{Sec.}{Secs.}
\Crefname{section}{Section}{main}
\Crefname{table}{Table}{Tables}
\crefname{table}{Tab.}{Tabs.}
\Crefname{figure}{Figure}{Figures}
\crefname{figure}{Fig.}{Figs.}

\def\ours{\texttt{\textbf{SViTT}}}
\def\cls{\textrm{cls}}

\DeclarePairedDelimiter\inner{\langle}{\rangle}
\DeclarePairedDelimiter\abs{\lvert}{\rvert}

\usepackage[font=small,skip=6pt]{caption}

\makeatletter
\renewcommand{\cvprsection}{\@startsection {section}{1}{\z@}
   {10pt plus 2pt minus 4pt}{4pt} {\large\bf}}
\renewcommand{\cvprsubsection}{\@startsection {subsection}{2}{\z@}
   {10pt plus 2pt minus 3pt}{3pt} {\elvbf}}
\renewcommand{\paragraph}{%
  \@startsection{paragraph}{4}%
  {\z@}{0.9ex \@plus 0.3ex \@minus .5ex}{-1em}%
  {\normalfont\normalsize\bfseries}%
}
\renewcommand\AB@affilsepx{\qquad \protect\Affilfont}
\makeatother

\input{math_commands.tex}

\def\cvprPaperID{10850} %
\def\confName{CVPR}
\def\confYear{2023}

\begin{document}

\title{\ours: Temporal Learning of Sparse Video-Text Transformers\vspace{-1em}}

\author[1]{Yi Li\thanks{Work done during an internship at Intel Labs.}}
\author[2]{Kyle Min}
\author[2]{Subarna Tripathi}
\author[1]{Nuno Vasconcelos}
\affil[1]{UC San Diego}
\affil[2]{Intel Labs
\authorcr\texttt{\small
\{yil898, nvasconcelos\}@ucsd.edu \qquad
\{kyle.min, subarna.tripathi\}@intel.com
}\vspace{-1em}
}

\maketitle

\input{main/0_abstract.tex}

\input{main/1_intro.tex}
\input{main/2_related-work.tex}
\input{main/3_sparse-vit.tex}
\input{main/4_expansion.tex}
\input{main/5_results.tex}
\input{main/6_conclusion.tex}

\paragraph{Acknowledgements.}
This work was funded in part by NSF award IIS-2041009.

{\small
\bibliographystyle{ieee_fullname}
\bibliography{main/ref}
}

\vspace{1em}
\appendix
\input{supp/1_implementation.tex}
\input{supp/2_results.tex}
\input{supp/3_discussions.tex}

\end{document}

%% file: math_commands.tex
\usepackage{amsmath,amsfonts,bm}

\def\eqref#1{equation~\ref{#1}}

\def\ceil#1{\lceil #1 \rceil}
\def\floor#1{\lfloor #1 \rfloor}
\def\1{\bm{1}}

\def\rva{{\mathbf{a}}}

\def\rvk{{\mathbf{k}}}

\def\rvp{{\mathbf{p}}}
\def\rvq{{\mathbf{q}}}

\def\rvv{{\mathbf{v}}}

\def\rvx{{\mathbf{x}}}
\def\rvy{{\mathbf{y}}}
\def\rvz{{\mathbf{z}}}

\def\rmB{{\mathbf{B}}}

\def\rmK{{\mathbf{K}}}

\def\rmP{{\mathbf{P}}}
\def\rmQ{{\mathbf{Q}}}
\def\rmR{{\mathbf{R}}}

\def\rmV{{\mathbf{V}}}
\def\rmW{{\mathbf{W}}}
\def\rmX{{\mathbf{X}}}
\def\rmY{{\mathbf{Y}}}
\def\rmZ{{\mathbf{Z}}}

\DeclareMathAlphabet{\mathsfit}{\encodingdefault}{\sfdefault}{m}{sl}
\SetMathAlphabet{\mathsfit}{bold}{\encodingdefault}{\sfdefault}{bx}{n}

\def\gA{{\mathcal{A}}}

\def\gE{{\mathcal{E}}}
\def\gF{{\mathcal{F}}}
\def\gG{{\mathcal{G}}}

\def\gJ{{\mathcal{J}}}

\def\gL{{\mathcal{L}}}
\def\gM{{\mathcal{M}}}
\def\gN{{\mathcal{N}}}

\def\gS{{\mathcal{S}}}

\def\gV{{\mathcal{V}}}

\newcommand{\R}{\mathbb{R}}

%% file: main/0_abstract.tex
\begin{abstract}
Do video-text transformers learn to model temporal relationships across frames? Despite their immense capacity and the abundance of multimodal training data, recent work has revealed the strong tendency of video-text models towards frame-based spatial representations, while temporal reasoning remains largely unsolved. In this work, we identify several key challenges in temporal learning of video-text transformers: the spatiotemporal trade-off from limited network size; the curse of dimensionality for multi-frame modeling; and the diminishing returns of semantic information by extending clip length. Guided by these findings, we propose \ours{}, a sparse video-text architecture that performs multi-frame reasoning with significantly lower cost than na\"ive transformers with dense attention. Analogous to graph-based networks, \ours{} employs two forms of sparsity: edge sparsity that limits the query-key communications between tokens in self-attention, and node sparsity that discards uninformative visual tokens. Trained with a curriculum which increases model sparsity with the clip length, \ours{} outperforms dense transformer baselines on multiple video-text retrieval and question answering benchmarks, with a fraction of computational cost.
Project page: \url{http://svcl.ucsd.edu/projects/svitt}.
\end{abstract}

%% file: main/1_intro.tex
\vspace{-4mm}
\section{Introduction}
\label{sec:1_intro}

\input{assets/figure/teaser.tex}

With the rapid development of deep neural networks for computer vision and natural language processing, there has been growing interest in learning correspondences across the visual and text modalities. A variety of vision-language pretraining frameworks have been proposed~\cite{li2019visualbert,lu2019vilbert,chen2020uniter,jia2021scaling} for learning high-quality cross-modal representations with weak supervision. Recently, progress on visual transformers (ViT)~\cite{dosovitskiy2021image,liu2021swin,bao2022beit} has enabled seamless integration of both modalities into a unified attention model, leading to image-text transformer architectures that achieve state-the-art performance on vision-language benchmarks~\cite{li2021align,singh2022flava,alayrac2022flamingo}.

Progress has also occurred in \emph{video}-language pretraining by leveraging image-text models for improved frame-based reasoning~\cite{bain2021frozen,fu2021violet,buch2022revisiting}. Spatial modeling has the advantage of efficient (linear) scaling to long duration videos. Perhaps due to this, single-frame models have proven surprisingly effective at video-text tasks, matching or exceeding prior arts with complex temporal components~\cite{buch2022revisiting,lei2022revealing}. However, spatial modeling creates a bias towards static appearance and overlooks the importance of temporal reasoning in videos. This suggests the question: Are temporal dynamics not worth modeling in the video-language domain?

Upon a closer investigation, we identify a few key challenges to incorporating multi-frame reasoning in video-language models. 
First, limited model size implies a trade-off between spatial and temporal learning (a classic example being 2D/3D convolutions in video CNNs~\cite{tran2018closer}). 
For any given dataset, optimal performance requires a careful balance between the two.
Second, long-term video models typically have larger model sizes and are more prone to overfitting.  Hence, for longer term video models, it becomes more important to carefully allocate parameters and control model growth.
Finally, even if extending the clip length improves the results, it is subject to diminishing returns since the amount of information provided by a video clip does not grow linearly with its sampling rate.  If the model size is not controlled, the computational increase may not justify the gains in accuracy. This is critical for transformer-based architectures, since self-attention mechanisms have a quadratic memory and time cost with respect to input length. In summary, model complexity should be adjusted adaptively, depending on the input videos, to achieve the best trade-off between spatial representation, temporal representation, overfitting potential, and complexity. Since existing video-text models
lack this ability, they either attain a suboptimal balance between spatial and temporal modeling, or do not learn meaningful temporal representations at all.  

Motivated by these findings, we argue that video-text models should learn to allocate modeling resources to the video data.  We hypothesize that, rather than uniformly extending the model to longer clips, the allocation of these resources to the relevant spatiotemporal locations of the video is crucial for efficient learning from long clips. For transformer models, this allocation is naturally performed by pruning redundant attention connections.
We then propose to accomplish these goals by exploring transformer sparsification techniques. This motivates the introduction of a {\it Sparse Video-Text Transformer\/} (\ours{}) inspired by graph models. As illustrated in \cref{fig:teaser}, \ours{} treats video tokens as graph vertices, and self-attention patterns as edges that connect them. We design \ours{} to pursue sparsity for both: 
\emph{edge} sparsity aims at reducing query-key pairs in attention module while maintaining its global reasoning capability; \emph{node} sparsity reduces to identifying informative tokens (e.g., corresponding to moving objects or person in the foreground) and pruning background feature embeddings.
To address the diminishing returns for longer input clips, we propose to train \ours{} with \emph{temporal sparse expansion}, a curriculum learning strategy that increases clip length and model sparsity, in sync, at each training stage. 

\ours{} is evaluated on diverse video-text benchmarks from video retrieval to question answering, comparing to prior arts and our own dense modeling baselines. First, we perform a series of ablation studies to understand the benefit of sparse modeling in transformers. Interestingly, we find that both nodes (tokens) and edges (attention) can be pruned drastically at inference, with a small impact on test performance. In fact, token selection using cross-modal attention improves retrieval results by 1\% without re-training. 

We next perform full pre-training with the sparse models and evaluate their downstream performance. 
We observe that \ours{} scales well to longer input clips where the accuracy of dense transformers drops due to optimization difficulties.
On all video-text benchmarks, \ours{} reports comparable or better performance than their dense counterparts with lower computational cost, outperforming prior arts including those trained with additional image-text corpora. 

The key contributions of this work are: 1) a video-text architecture \ours{} that unifies edge and node sparsity; 2) a sparse expansion curriculum for training \ours{} on long video clips; and 3) empirical results that demonstrate its temporal modeling efficacy on video-language tasks.

%% file: assets/figure/teaser.tex
\begin{figure}[t]
    \centering
    \includegraphics[width=\linewidth]{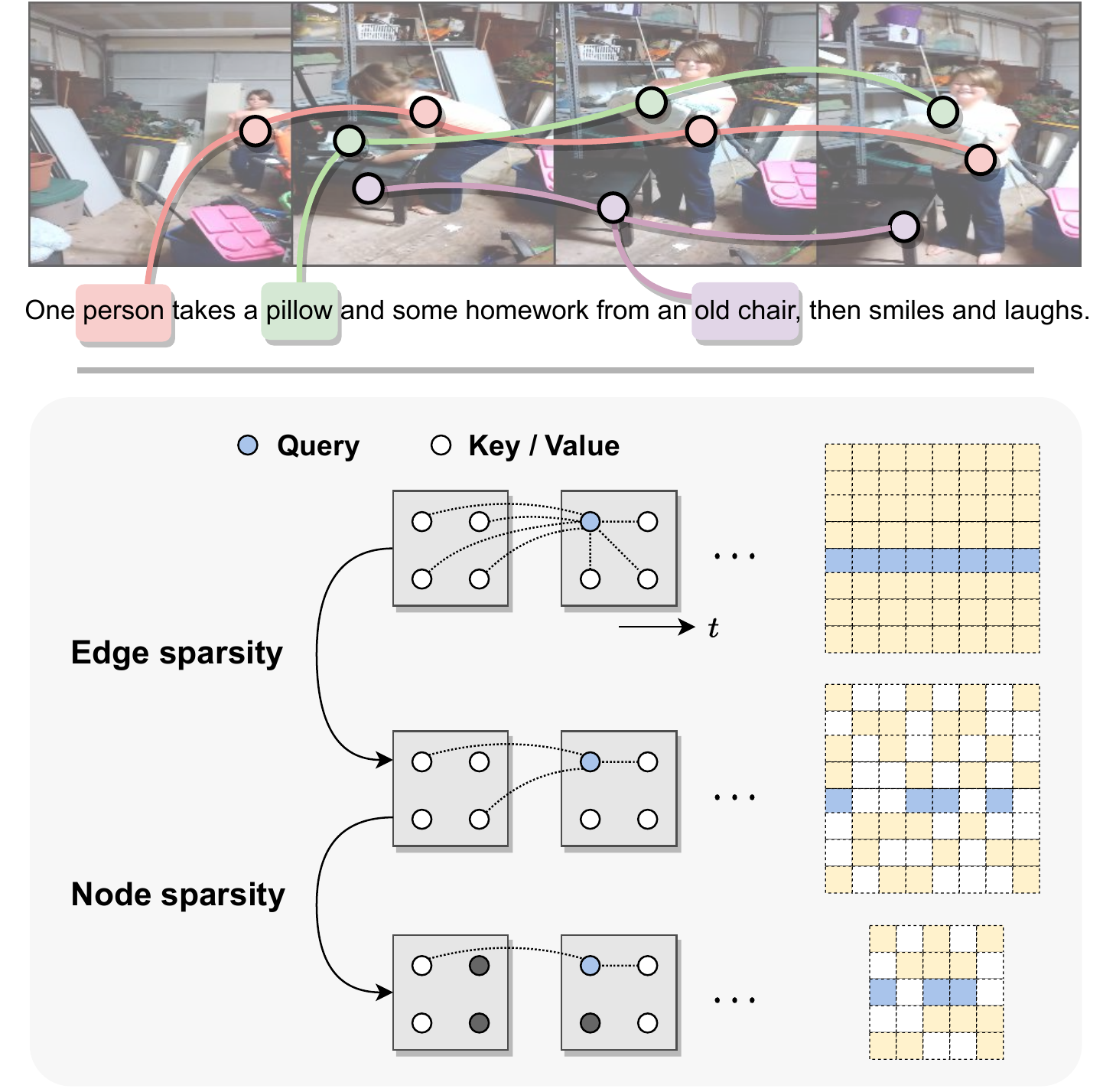}
    \caption{We propose \ours{}, a \emph{sparse} video-text transformer for efficient modeling of temporal relationships across video frames. \textbf{Top}: Semantic information for video-text reasoning is highly localized in the spatiotemporal volume, making dense modeling inefficient and prone to contextual noises. \textbf{Bottom}: \ours{} pursues \emph{edge} sparsity by limiting query-key pairs in self-attention, and \emph{node} sparsity by pruning redundant tokens from visual sequence.}
    \label{fig:teaser}
\end{figure}

%% file: main/2_related-work.tex
\section{Related Work}
\label{sec:2_related-work}

\paragraph{Video-language pretraining.}
Vision-language pretraining has been widely adopted for various video-text downstream tasks. VideoBERT~\cite{sun2019videobert} was an early effort, using video-text pretraining for action classification and video captioning. Recently, the massive-scale instructional video dataset HowTo100M~\cite{miech2019howto100m} has motivated many approaches to video-text pretraining~\cite{miech2020end,gabeur2020multi,zhu2020actbert,zellers2021merlot,bain2021frozen,lei2022revealing}. 
Frozen~\cite{bain2021frozen} proposed to pretrain a space-time transformer on a combination of video and image data to enable zero-shot text-to-video retrieval.
ATP~\cite{buch2022revisiting} and Singularity~\cite{lei2022revealing} showed strong performance using image-based models, highlighting the importance of spatial modeling for video-language tasks.
In this work, we pursue an alternative route of efficient \emph{temporal} modeling across multiple video frames.

\paragraph{Sparse transformers.}
The self-attention of na\"ive transformers~\cite{vaswani2017attention,dosovitskiy2021image} has quadratic complexity making them inefficient for modeling long sequences.
Different forms of sparse attention have been studied to improve text~\cite{child2019generating,beltagy2020longformer,zaheer2020big}, image~\cite{vaswani2021scaling,liu2021swin,dong2022cswin}, and video modeling~\cite{bertasius2021space,arnab2021vivit,liu2022video}, although the sparse patterns are typically predetermined and do not adapt to the input semantics.
Several works have also considered speeding up vision transformers by adaptively reducing the number of input tokens~\cite{rao2021dynamicvit,chen2021chasing,NEURIPS2021_6a30e32e,liang2022evit,yin2022vit,meng2022adavit}. For example,
DynamicViT~\cite{rao2021dynamicvit} proposed to drop visual tokens with a dedicated module that identifies and prunes less informative ones. 
TokenLearner~\cite{NEURIPS2021_6a30e32e} introduces a learnable module to adaptively generate a small subset of tokens from input frames. 
EViT~\cite{liang2022evit} progressively reduces the number of tokens based on their attention scores, fusing inattentive tokens into a new token to preserve input information.
Unlike prior works that focus on visual modeling on images or videos alone, we study the sparsity of video-language transformers which can benefit from cross-modal attention. 

%% file: main/3_sparse-vit.tex
\section{Exploiting Sparsity in Video Transformers}
\label{sec:3_sparse-vit}

\input{assets/figure/arch.tex}

In this section, we formulate video transformers as graph models (\cref{sec:3.1_prelim}) and present a set of approaches towards sparse video modeling, exploiting the redundancy of edges (\cref{sec:3.2_edge}) and nodes (\cref{sec:3.3_node}). We combine these into a unified sparse framework for video-text learning (\cref{sec:3.4_hybrid}).

\subsection{Video Transformers are Graph Models}
\label{sec:3.1_prelim}

Visual transformers~\cite{dosovitskiy2021image} are deep neural networks that model images or videos as sequences of local pixel patches, through a combination of patchwise feature transformation and self-attention. Inspired by transformer architectures for language models~\cite{vaswani2017attention}, video transformers encode input clips into a sequence of spatiotemporal patches, flattened and linearly projected to a $d$-dimensional embedding space:
\begin{equation}
    \rmZ^{(0)} = \left(\rvz_\cls^{(0)}, \rvz_1^{(0)}, \dots, \rvz_N^{(0)}\right) = f^{\textrm{tok}}(\rvx_{1:T}) \in \R^{(N + 1) \times d}
\end{equation}
where $N = T' H' W'$ is the volume of the 3D patch grid,
and $\rvz_\cls^{(0)}$ denotes a special class token responsible for instance-level prediction. The tokenized sequence is processed by a cascade of transformer blocks $f^{(l)}$
\begin{equation}
    \rmZ^{(l)} = f^{(l)}(\rmZ^{(l-1)}), \quad l = 1, \dots, L
\end{equation}
each of which computes the self-attention $\gA$ between input tokens, followed by a feed-forward network $\gF$:\footnote{Attention heads, residual connections and normalization terms are omitted for brevity, although we use the conventional implementation~\cite{vaswani2017attention}.}
\begin{gather}
    f^{(l)}(\rmZ) = \gF(\gA(\rmZ \rmW_K^T, \rmZ \rmW_Q^T, \rmZ \rmW_V^T)), \\[1ex]
    \gA(\rmQ, \rmK, \rmV) = \sigma(\rmQ \rmK^T) \rmV \label{eqn:self-attn}
\end{gather}

We interpret the transformer architecture as a special case of \emph{graph} networks~\cite{battaglia2018relational}, with \emph{nodes} representing tokenized video patches and \emph{edges} connecting pairs of tokens for which self-attention is computed. 
Specifically, consider a directed graph $\gG = (\gV, \gE)$ defined by the vertices $\gV = \{\rvz_1, \dots, \rvz_N\}$ corresponding to spatiotemporal video patches, and edges $\gE \subseteq \{1, \dots, N\}^2$ connecting pairs of nodes. The self-attention of (\ref{eqn:self-attn}) can be generalized such that node $\rvz_i$ attends to $\rvz_j$ only if $(i, j) \in \gE$, 
\begin{gather}
    \gA_\gE(\rvq_i, \rmK, \rmV) = \sum_{j: (i, j) \in \gE} \rva_{ij} \rvv_j, \label{eq:ga}\\[1ex]
    \rva_{ij} = \frac{e^{\inner{\rvq_i, \rvk_j}}}{\sum_{j: (i, j) \in \gE} e^{\inner{\rvq_i, \rvk_j}}} 
\end{gather}

Under this interpretation, the transformer architecture with full attention resembles a \emph{complete} graph, where $\gE$ includes every pair of vertices. This dense attention mechanism endows (\ref{eq:ga}) with quadratic memory and time complexity w.r.t. sequence length $N$, making na\"ive transformers notoriously inefficient to train and expensive to deploy, especially for longer video clips. We argue that due to the inherent sparsity of information in video data, a large portion of the graph can be pruned dynamically without significant performance loss, leading to a \emph{sparse} graph model of significantly lower cost for training and inference. 

\subsection{Edge Sparsity: Local \& Random Attention}
\label{sec:3.2_edge}

Prior natural language processing models, such as BigBird~\cite{zaheer2020big}, have explored the idea of restricting the number of key-value pairs each query token attends to, which reduced the number of edges in $\gE$.
We use a similar procedure to create a video transformer with \emph{edge sparsity}, utilizing a combination of \emph{local}, \emph{random} and \emph{global} attention.

\paragraph{Local attention.}
Regional tokens $\{\rvz_i\}_{i=1}^N$ are first chunked into $N_b = \ceil{N/G}$ contiguous blocks of size $G$\footnote{Padding is added when $N$ is not divisible by $G$.}. Tokens of one block $k$ attend to a local neighborhood of $K_l$ blocks $\{k - \Delta, \dots, k + \Delta\}$, where $\Delta = (K_l - 1) / 2$ is the maximum range of local attention,
\begin{equation}
    (k, k') \in \gE, \quad \forall k, k': \abs{k' - k} \le \Delta
    \label{eqn:local-edge}
\end{equation}
This preserves the modeling of interactions between local features (objects, people, textures) that does not require long-range attention. In the  case of $K_l = 1$, local reduces to diagonal attention, where each block only attends to itself.

\paragraph{Random attention.}
Beyond \emph{local} attention, each block also attends to $K_r$ other blocks sampled randomly from the input sequence,
\begin{equation}
    (k, k') \in \gE, \quad \forall k' \in \gN(k)
    \label{eqn:random-edge}
\end{equation}
where $\gN(k)$ is a random subset of $\{k' \mid \abs{k' - k} > \Delta\}$ of size $K_r$.
This enables the transformer to model long-range visual relationships while avoiding the quadratic cost. 

\paragraph{Global attention.}
Class token $\rvz_\cls$ always attends to/from \emph{regional} tokens $\rvz_i$, i.e. the link between $\rvq_\cls$ and $\rvk_i$, as well as $\rvq_i$ and $\rvk_\cls$, are always retained:
\begin{equation}
    (\cls, i), (i, \cls) \in \gE, \quad i = 1, \dots, N
    \label{eqn:global-edge}
\end{equation}
This enables $\rvz_\cls$ to capture global video context, even when the rest of tokens do not attend globally.

In summary, we retain all attention connections for local-to-global and global-to-local vertices, and limit local-to-local edges to $(K_l + K_r) G$ per token. Edge sparsity has a \emph{linear} asymptotic cost of $O((K_l + K_r) GN)$, a significant improvement over dense attention at $O(N^2)$. However, this approach has two critical limitations of this strategy. First, the sparse patterns are predetermined and do not adapt dynamically to the input sequence. Second, the connections that remain in $\gE$ are not determined by the video semantics. In result, connections between pairs of tokens of low semantic affinity (low attention values) may be preserved, impairing the efficiency of the sparse attention mechanism. To enable more aggressive sparsification, we introduce a second mechanism, which is dynamic, guided by video semantics, and applied to the \emph{nodes} of the graph.

\subsection{Node Sparsity: Dynamic Token Pruning}
\label{sec:3.3_node}

A large percentage of the tokens of a video transformer corresponds to \emph{contextual} regions, which contain little temporal dynamics and are only weakly related to the prediction target (e.g. background content uninformative of the activities of subjects in the foreground). While edge sparsification improves the efficiency of self-attention, it lacks both the flexibility and the semantic sensitivity to account for the uneven distribution of information across video patches.

To introduce these properties, we propose a node sparsification strategy, based on the dynamic pruning of tokens. We leverage a combination of observations. First, video semantics are summarized by the class tokens $\rvz_\cls$, which contain a global representation of the information of interest for classification. Second the global-to-local edges survive the edge sparsification, through (\ref{eqn:global-edge}), making the attention weights from the $\rvz_\cls$ to all regional tokens $\rvz_i$,
\begin{equation}
    \rva(\rvz_i) = \frac{e^{\inner{\rvq_\cls, \rvk_i}}}{\sum_{j=1}^N e^{\inner{\rvq_\cls, \rvk_j}}}, \quad i = 1, \dots, N
    \label{eqn:cls-attn}
\end{equation}
available for node sparsification. Since the class token is used for video-level predictions, $\rva(\rvz_i)$ quantifies the contribution of feature $\rvz_i$ to the main task. This implies that nodes $\rvz_i$ of low $\rva(\rvz_i)$ are not informative and can be ignored~\cite{liang2022evit}.

 Node pruning then reduces to keeping the $N' = \ceil{q N}$ tokens of largest class attention
\begin{equation}
    \gS_\gV(\rmZ; q) = \{\rvz_i \mid i \in \mathrm{topk}(\rva(\rmZ), \ceil{q N})\}
    \label{eqn:node-sparse}
\end{equation}
where hyperparameter $q$ denotes \emph{keep rate} and $\mathrm{topk}({\bf L}, m)$ selects the $m$ largest entries of vector $\bf L$. 
This procedure is repeated multiple times throughout the video encoder (keep rate $q^{(l)}$ at layer $l$), progressively reducing the length of input sequences. Importantly, the pruning procedure is dynamic and ensures that semantically uninformative vertices in the attention graph $\gG$ are removed along with all edges they are associated with. 

\paragraph{Cross-modal sparsity.}
Node sparsification can be naturally extended to video-language learning. For this, we propose to extend the token selection mechanism discussed above to a \emph{cross-modal} setting, where video and text tokens $\rmZ_v, \rmZ_t$ are modeled jointly in a multimodal encoder. In this case, we replace the query $\rvq_\cls$ of~(\ref{eqn:cls-attn}) with the class token of text sequence $\rvq_\cls^{(t)}$, obtaining cross-modal attention
\begin{equation}
    \rva_m(\rvz_i^{(v)}) = \frac{e^{\inner{\rvq_\cls^{(t)}, \rvk_i^{(v)}}}}{\sum_{j=1}^N e^{\inner{\rvq_\cls^{(t)}, \rvk_j^{(v)}}}}, \quad i = 1, \dots, N
    \label{eqn:cross-attn}
\end{equation}
and subsequently the token sparsification function
\begin{equation}
    \gS_\gM(\rmZ_v; q) = \{\rvz_i^{(v)} \mid i \in \mathrm{topk}(\rva_m(\rmZ_v), \ceil{q N})\}
    \label{eqn:xmodal}
\end{equation}

Multimodal node sparsity $\gS_\gM$ is applied on top of visual sparsity $\gS_\gV$. We expect cross-modal sparsification to create additional room for sparsity over standalone visual modeling. While the visual encoder can identify salient actors and objects from background patches from the input clip, only the text semantics provide direct guidance for the video-text model to focus on regions relevant to the task of interest. 

With node sparsity, the compute cost of subsequent layers is improved to $O(q^2 N^2)$ using dense attention or $O(q(K_l + K_r) GN)$ with edge sparsity, and the reduction accumulates with multiple sparse layers. 

\subsection{Hybrid Sparse Transformers}
\label{sec:3.4_hybrid}

We propose to combine \emph{edge} and \emph{node} sparsity into a unified framework, \ours{}, as illustrated in~\cref{fig:sparse-vit}. \ours{} is built on top of existing video-language transformer architectures that combine separate video and text encoders with a cross-modal transformer. 
The visual encoder blocks perform sparse self-attention in the following steps:
\begin{itemize}
    \item \emph{edge sparsification}: given the random graph $\gE$ of (\ref{eqn:local-edge})-(\ref{eqn:global-edge}) compute attention weights  $\gA_\gE$, using (\ref{eq:ga}); 
    \item \emph{node sparsification:} select the nodes $\gS_\gV$ to retain, using  (\ref{eqn:cls-attn})-(\ref{eqn:node-sparse}) with keep rate $q < 1$).
\end{itemize}

After video and text embeddings $\rvz_v, \rvz_t$ are derived from their respective encoders, a multimodal transformer is applied to aggregate features across modalities. It follows the design of the text encoder, except for a cross-attention module that is applied after each self-attention block, where text queries $\rvq_t$ attends to key-value pairs $\rvk_v, \rvv_v$ extracted by the video encoder. The text-to-video attention of (\ref{eqn:cross-attn})-(\ref{eqn:xmodal}) is then used to select the nodes $\gS_\gM$ to retain, further reducing the number of video tokens for subsequent layers. 

%% file: assets/figure/arch.tex
\begin{figure*}[t]
    \centering
    \includegraphics[width=\linewidth]{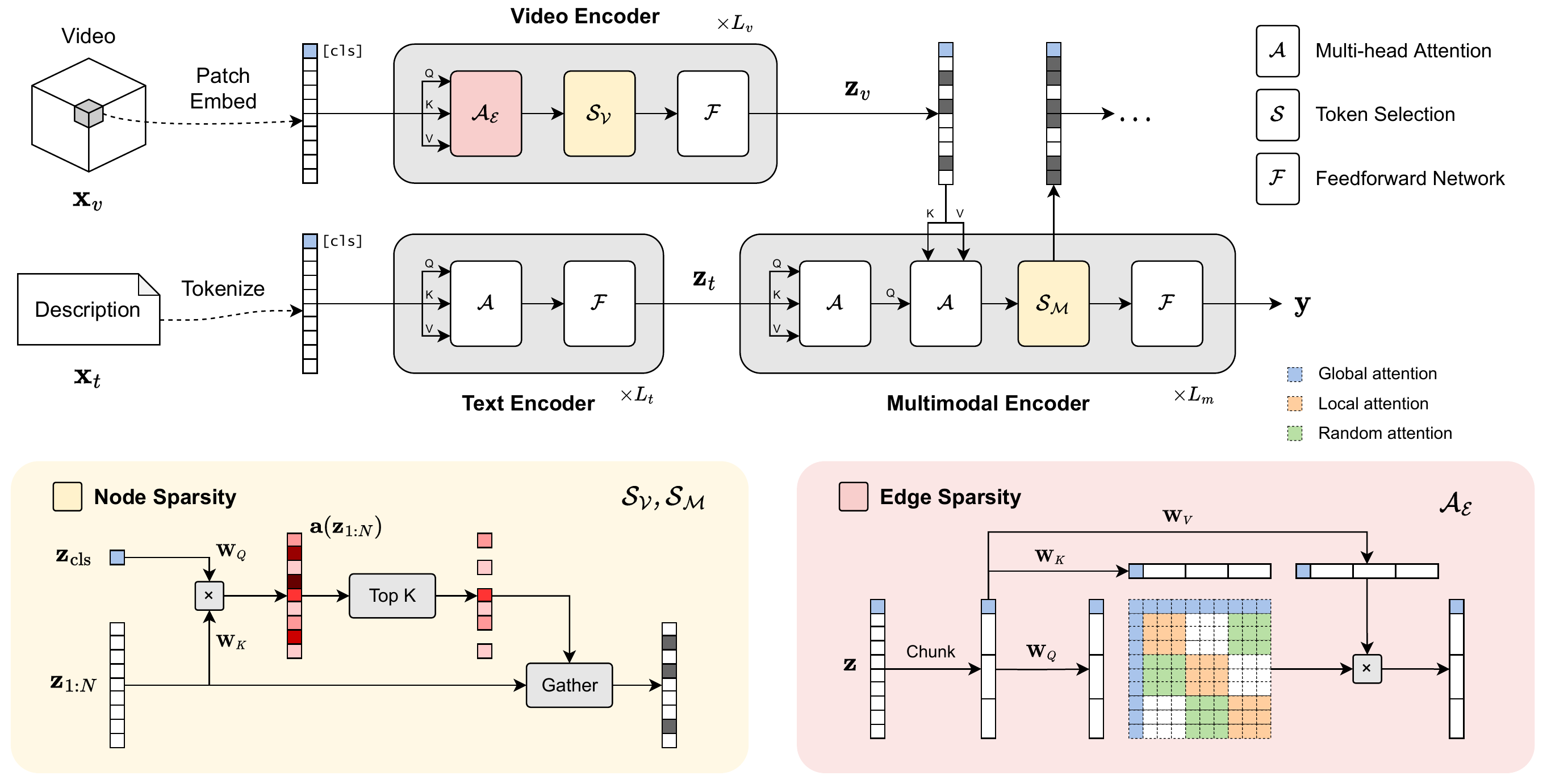}
    \caption{\textbf{Model Architecture.} \ours\ improves modeling efficiency of conventional video-text transformers through two key components: \emph{node} sparsity and \emph{edge} sparsity. \textbf{Edge sparsification} $\gA_\gE$ computes sparse self-attention of input visual sequence $\rvz$, where each query token attends to a small subset of key and value tokens, with connectivity $\gE$ specified by global, local, and random attention. \textbf{Node sparsification} $\gS_\gV$ uses global attention scores from $\gA_\gE$ to prune uninformative tokens, removing them from the computational graph of subsequent layers; $\gS_\gM$ uses text-to-video attention in the multimodal encoder to further reduce the length of visual sequence.}
    \label{fig:sparse-vit}
\end{figure*}

%% file: main/4_expansion.tex
\section{Temporal Sparse Expansion}
\label{sec:4_expansion}

In this section, we introduce a new training strategy for \ours{}.  We motivate for progressive model training with increasing clip length and sparsity in~\cref{sec:4.1_sparsity}, and detail our pretraining procedure in~\cref{sec:4.2_expansion}.

\subsection{Sparsity vs. Clip Length}
\label{sec:4.1_sparsity}

The key insight behind the design of \ours{} is the \emph{diminishing return} of clip length. In general, a $2\times$ longer sequence does not contain twice the semantic information about the video, due to the redundancy of adjacent frames. This leads to a lower percentage of informative patches with denser frame sampling. 

Due to this redundancy, it is possible to use higher sparsity for models with longer clips. This can be implemented by reducing the keep rate $q$ of node sparsification, and the percentage of key/value blocks to attend to in edge sparsification ($K_l/N_b, K_r/N_b$). For the latter, since the total number of blocks $N_b = \ceil{N/G}$ increases with number of frames $T$ (assuming a fixed block size $G$), it suffices to keep the parameters $K_l, K_r$ constant.

\subsection{Temporal Expansion}
\label{sec:4.2_expansion}

\input{assets/figure/expansion.tex}

Pretraining video-text transformers on long clips is time-consuming and leads to suboptimal models. Instead, we follow a learning strategy similar to Frozen~\cite{bain2021frozen}, where the model is initially pretrained  with shorter clips, and the number of frames increases as training progresses. However, when expanding the clip length, we increase the model sparsity to simultaneously 1) account for the redundancy of information, and 2) limit the growth of computational cost. 

\cref{fig:prog-expansion} depicts the expansion process proposed for video-text training. In the initial training stage $j = 1$, a dense video-text model is pretrained on clip length $T_1$. Denoting the sparsity hyperparameters of \ours{} at stage $j$ by $S_j = (q_j, K_l^{(j)}, K_r^{(j)})$, we create a learning curriculum with progressively increasing clip length and sparsity, by enforcing the constraints
\begin{gather}
    T_1 < T_2 < \dots; \\
    q_1 > q_2 > \dots; \label{eq:Qchain} \\
    \frac{K_l^{(1)} + K_r^{(1)}}{T_1} > \frac{K_l^{(2)} + K_r^{(2)}}{T_2} > \dots \label{eq:Kchain}
\end{gather}
In practice, we use a decreasing token keep rate (\ref{eq:Qchain}) and keep local and random attention block numbers fixed, i.e. $K_l^{(j)} = K_l$ and $K_r^{(j)} = K_r$, to satisfy (\ref{eq:Kchain}).

%% file: assets/figure/expansion.tex
\begin{figure}[t]
    \centering
    \includegraphics[width=\linewidth]{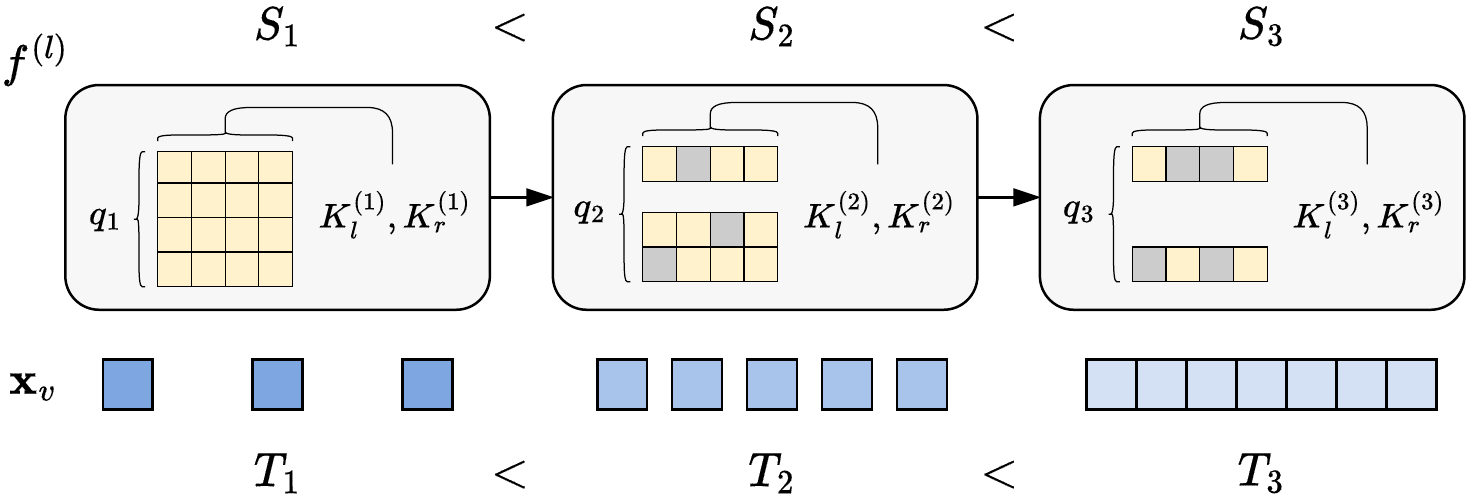}
    \caption{\textbf{Temporal Sparse Expansion.} We propose a multi-stage curriculum for training \ours. At each stage, the node and edge sparsity $S$ of video-text transformer increases with clip length $T$.}
    \label{fig:prog-expansion}
\end{figure}

%% file: main/5_results.tex
\section{Results}
\label{sec:5_results}

In this section we present experimental results of \ours{} on vision-language modeling. We briefly introduce the experimental setup in \cref{sec:5.1_setup}, and perform several ablation studies on the design choices involving model sparsification and training in \cref{sec:5.2_ablation}. We then demonstrate the performance of \ours{} on various vision-language tasks in \cref{sec:5.3_main-result} and include additional qualitative analysis.

\subsection{Experimental Setup}
\label{sec:5.1_setup}

\input{assets/table/sparsity-ablation.tex}
\input{assets/figure/sparse-block-attn.tex}

\paragraph{Architecture.}
Our implementation of video-text transformer is based on \emph{Singularity}~\cite{lei2022revealing}. The model has a two-tower structure, with separate encoders for vision and language. The video encoder $f_v$ is a 12-layer BEiT-B~\cite{bao2022beit} initialized with ImageNet weights and inflated for video inputs. This differs from~\cite{lei2022revealing} which embeds each frame independently and applies late temporal fusion on extracted features. The text encoder $f_t$ is a pretrained BERT~\cite{devlin2018bert} model, whose last 3 layers are modified to implement the multimodal encoder $f_m$, with \emph{cross-modal} attention based on visual tokens as key-value pairs, which we sparsify as described in \cref{sec:3.3_node}.

\paragraph{Pre-training.}
\ours{} is pre-trained on 2.5M video-text pairs from the WebVid dataset~\cite{bain2021frozen}. Since our goal is to investigate how to improve the effectiveness of the \emph{temporal} learning of the video modality, we do not train with additional image-text corpora as done in~\cite{bain2021frozen,fu2021violet,lei2022revealing}. 

The \texttt{[cls]} tokens  of the video  and text encoders are first linearly projected  to a joint embedding space, producing feature vectors  $\rvz_v = f_v(\rvx_v)$  and $\rvz_t = f_t(\rvx_t)$, respectively. Following prior work, we use the InfoNCE loss~\cite{oord2018representation}
to align these feature vectors. 
The output of multimodal encoder $\rvy = f_m(\rvz_v, \rvz_t)$ is optimized with video-text matching (VTM)
and masked language modeling (MLM) losses
commonly found in VLP literature\cite{chen2020uniter,li2020hero,li2021align,fu2021violet,lei2022revealing}. 

\paragraph{Downstream tasks.}
Trained video-text models are evaluated on two multimodal tasks: \emph{text-to-video retrieval} and \emph{video question answering}. 
Video retrieval is evaluated on MSR-VTT~\cite{xu2016msr}, DiDeMo~\cite{anne2017localizing}, Charades~\cite{sigurdsson2016hollywood} and Something-Something v2~\cite{goyal2017something,lei2022revealing}, by top-$K$ recalls ($K \in \{1, 5, 10\}$) and their numeric average. Question answering is evaluated on MSRVTT-QA~\cite{xu2017video}, ActivityNet-QA~\cite{caba2015activitynet,yu2019activitynet} and AGQA~\cite{grunde2021agqa}, with top-1 accuracy of the answers.

Training details are given in the Appendix.

\subsection{Ablation Studies}
\label{sec:5.2_ablation}

\input{assets/table/sparse-model-ablation.tex}
\input{assets/figure/node-keep-rate.tex}

We start by training a video-text transformer with clip length of 4 frames and \emph{dense} attention, and measure its zero-shot performance on downstream tasks after \emph{sparsifying} its video encoder while keeping its weights unchanged. 

\paragraph{Edge sparsification.}
We first apply edge sparsification with different number of local blocks $K_l$, random blocks $K_r$, and block size $G$. 
As shown in \cref{fig:sparse-attn-blocks}, under a limited budget of 6 local or random attention blocks per query token, using 1 local block and 5 random blocks provides the best trade-off. 
Increasing the local attention window $K_l$ hurts long-term modeling capacity and degrades retrieval; while removing the single local block responsible for diagonal attention also impairs performance. 
This suggests that while query tokens should always attend to their respective blocks, there is no benefit in attending to neighboring blocks. We thus fix $K_l = 1$ for the rest of experiments.

We next vary the block size $G$ of sparse attention. \cref{fig:sparse-attn-block-size} shows that larger sizes have stronger retrieval performance, as expected. However, they also make self-attention less sparse (more costly to compute). $G = 56$ provides a good balance between performance and complexity.
\cref{tab:sparsity-ablation} summarizes the test performance of two edge sparsity configs: $(K_l, K_r, G) = (1, 3, 56)$ and $(1, 5, 56)$. While both underperform the dense model, we will later demonstrate that the gap can be closed and even reversed by training the sparse model with the proposed temporal expansion curriculum.

\input{assets/table/sparse-expansion.tex}

\paragraph{Node sparsification.}
We next incorporate node sparsity into the video-text transformer. This includes \emph{visual} sparsification (keep rate $q_v$) using the self-attention of video encoder $f_v$ to progressively prune the input tokens\footnote{We follow~\cite{liang2022evit} to prune visual tokens at layer \#4, \#7, and \#10.}, and \emph{multimodal} sparsification (keep rate $q_m$) using the text-to-video attention of cross-modal encoder $f_m$ to further drop visual tokens unrelated to the text query.
\cref{fig:visual-keep-rate} shows that the dense model is quite robust to token pruning, even without sparse training. Using visual keep rate $q_v \ge 0.8$ has minimal impact on test results, and performance only starts to drop rapidly at $q_v = 0.5$, at which point only $1/8$ of input tokens are retained after three rounds of sparsification.

Even more surprisingly, the subsequent multimodal sparsification step, using a keep rate of $q_m = 0.1$, \emph{improves} zero-shot performance by $1\%$. This shows that the $90\%$ redundant visual tokens not only add unnecessary complexity to the model, but also introduce noise that harms retrieval performance. The fact that the optimal $q_m$ is much lower than $q_v$ also suggests that text modality provides crucial semantic guidance for identifying relevant visual patches, at a much higher accuracy than visual modeling alone.

\input{assets/table/zs-retrieval_v2.tex}

\paragraph{Hybrid sparsification.}
Combining the best sparse settings for edges and nodes, we obtain the hybrid sparsification strategy for \ours. As shown in \cref{tab:sparsity-ablation}, compared to edge sparsity, the introduction of node sparsity ($q_v = 0.7, q_m = 0.1$) only impacts recall scores marginally, while saving computations on a large portion of visual tokens throughout the network. 

\input{assets/figure/qualitative.tex}

\paragraph{Training sparse transformers.}
We next perform \emph{full pretraining} with the sparse models and compare their performance and efficiency to the dense transformer baseline. \cref{tab:sparse-model-ablation} summarizes the results obtained with different input clip lengths and types of sparsity. At 4 frames,  edge sparsity has small benefit due to the relatively short sequence length and node sparsity performs the best. However, at 8 frames, we start to see clear advantages to edge sparsity in memory complexity, and both edge and node sparsity outperform the dense transformer baseline. Combining both types of sparsity performs the best, while only requiring 60\% the FLOPs and 30\% the memory of the dense model.
At 16 frames, the dense and node sparsity models are no longer trainable due to their quadratically increasing cost. The models with edge sparsity, however, are able to fit into GPU memory thanks to the linear complexity from sparse attention. A 16-frame \ours{} with hybrid sparsity requires similar computation to an 8-frame dense model and only half of its training memory, while achieving 3\% higher recall.

\paragraph{Progressive training.}
We next study the impact of temporal sparse expansion on the training of \ours{} models on longer clips. \cref{tab:sparse-expansion} compares the performance of models trained using temporal expansion (i.e. initialized from checkpoints pretrained on fewer frames and lower sparsity) to standard single-stage training. For single-stage training, performance does not improve substantially with clip length. This is in contrast to the proposed sparse expansion, where using $8$ instead or $4$ frames results in a gain of $2.7\%$. This suggests that the models have learned to exploit the temporal relationships between video frames. For a given clip length, sparse expansion also substantially improves upon single stage performance, from $2.8$ points for $T=4$ to $4.5$ points for $T = 16$. In fact, sparse expansion training with a shorter clip length (e.g. $4$ frames) can outperform single stage training with a larger length ($16$ frames).

\subsection{Main Results}
\label{sec:5.3_main-result}

\input{assets/table/ft-retrieval_v2.tex}

We compare \ours{} to state-of-the-art models in text-to-video retrieval and video question answering.

\paragraph{Text-to-video retrieval.}
Video-text retrieval is evaluated under zero-shot and fine-tuning settings.
\cref{tab:zs-retrieval} shows the zero-shot results on MSR-VTT and DiDeMo. 
Compared to our reproduced models of Singularity on WebVid-2M, which aggregate frame-level features using a temporal transformer encoder, the spatiotemporal transformer of \ours{} produces stronger results on both downstream datasets. This highlights the importance of temporal modeling in earlier layers of video-text transformers. 
\ours{} with hybrid sparsity has similar performance to the dense model on MSR-VTT but significantly outperforms in on DiDeMo, which contains longer videos with localized activities.

For Charades and SSv2, we evaluate text-to-video retrieval with fine-tuning, as shown in \cref{tab:ft-retrieval}. Both datasets are action-centric, posing a greater challenge to the temporal reasoning of video-text models. \ours{} with hybrid sparsification dominates the dense variant by $\sim 3\%$ on both datasets, a more substantial gap than observed for MSR-VTT and DiDeMo. This confirms our hypothesis that exploiting node and edge sparsity reduces the dependency of models on contextual regions, forcing them to focus on the temporal dynamics of person and objects in the foreground.

\input{assets/table/video-qa_v1.tex}

\paragraph{Video question answering.}
We next evaluate the cross-modal modeling of \ours{} on MSRVTT-QA, ActivityNet-QA and AGQA. As shown in \cref{tab:video-qa}, the hybrid sparsity version of \ours{} outperforms the dense transformer baseline on all three datasets, thanks to its more efficient temporal modeling. On MSRVTT-QA, the accuracy gap is small between dense and sparse transformer (0.3\%), and \ours{} marginally underperforms baselines pretrained with fewer number of frames (MERLOT~\cite{zellers2021merlot}, VIOLET~\cite{fu2021violet}). This is likely due to the nature of MSRVTT, which consists of short video clips and questions biased towards spatial cues, allowing temporal modeling little benefit over spatial transformers pretrained on massive image \& video data. On ActivityNet-QA and AGQA, which both contain longer clips and temporally challenging questions, the sparse modeling of \ours{} proves advantageous, with 0.9\% and 2.3\% boost in accuracy respectively, beating all baseline methods.

\paragraph{Qualitative analysis.}
To show how \ours{} efficiently identifies and concentrates its computation on informative spatiotemporal regions of the input clips, we visualize the outcome of node sparsification in \cref{fig:qualitative}. 
Using visual sparsification in video encoder $f_v$, \ours{} learns to isolate foreground entities from the majority of background patches, enabling the model to perform sparse video-text inference on longer temporal context. 
Cross-modal attention in multimodal encoder $f_m$ provides an even stronger signal for isolating the regions of interest of each video clip, validating the importance of text semantics in visual sparsification.

%% file: assets/table/sparsity-ablation.tex
\begin{table}[t]
    \centering

    \resizebox{\linewidth}{!}{
\begin{tabular}{@{}ccccccc@{}}
\toprule
\textbf{Attn. blocks} & \textbf{Keep rate} & \textbf{\# Edges} & \multicolumn{4}{c}{\textbf{DiDeMo}} \\
$K_l, K_r, G$ & $q_v, q_m$ & (M) & R1 & R5 & R10 & \textbf{Mean} \\ \midrule
--- & --- & 7.47 & 28.8 & 53.1 & 63.0 & \textbf{48.3} \\ \midrule
\multicolumn{7}{c}{\textit{Edge sparsity}} \\ \midrule
(1, 3, 56) & --- & 2.14 & 20.7 & 41.7 & 50.5 & \textbf{37.6} \\
(1, 5, 56) & --- & 3.21 & 26.0 & 48.6 & 56.8 & \textbf{43.8} \\ \midrule
\multicolumn{7}{c}{\textit{Node sparsity}} \\ \midrule
--- & (0.7, 1) & 3.99 & 26.9 & 51.9 & 61.3 & \textbf{46.7} \\
--- & (0.7, 0.1) & 3.97 & 27.6 & 53.1 & 62.9 & \textbf{47.9} \\ \midrule
\multicolumn{7}{c}{\textit{Hybrid sparsity}} \\ \midrule
(1, 3, 56) & (0.7, 0.1) & 1.48 & 19.9 & 40.5 & 50.6 & \textbf{37.0} \\
(1, 5, 56) & (0.7, 0.1) & 2.22 & 24.5 & 47.6 & 58.6 & \textbf{43.6} \\ \bottomrule
\end{tabular}
    }

    \caption{\textbf{Ablation on Edge and Node Sparsity}. We evaluate the same \emph{dense} video-text model under different sparsification modes. 
    }
    \label{tab:sparsity-ablation}
\end{table}

%% file: assets/figure/sparse-block-attn.tex
\begin{figure}[t]
    \centering
    
    \begin{subfigure}[b]{.49\linewidth}
    \centering
    \includegraphics[height=0.6\linewidth]{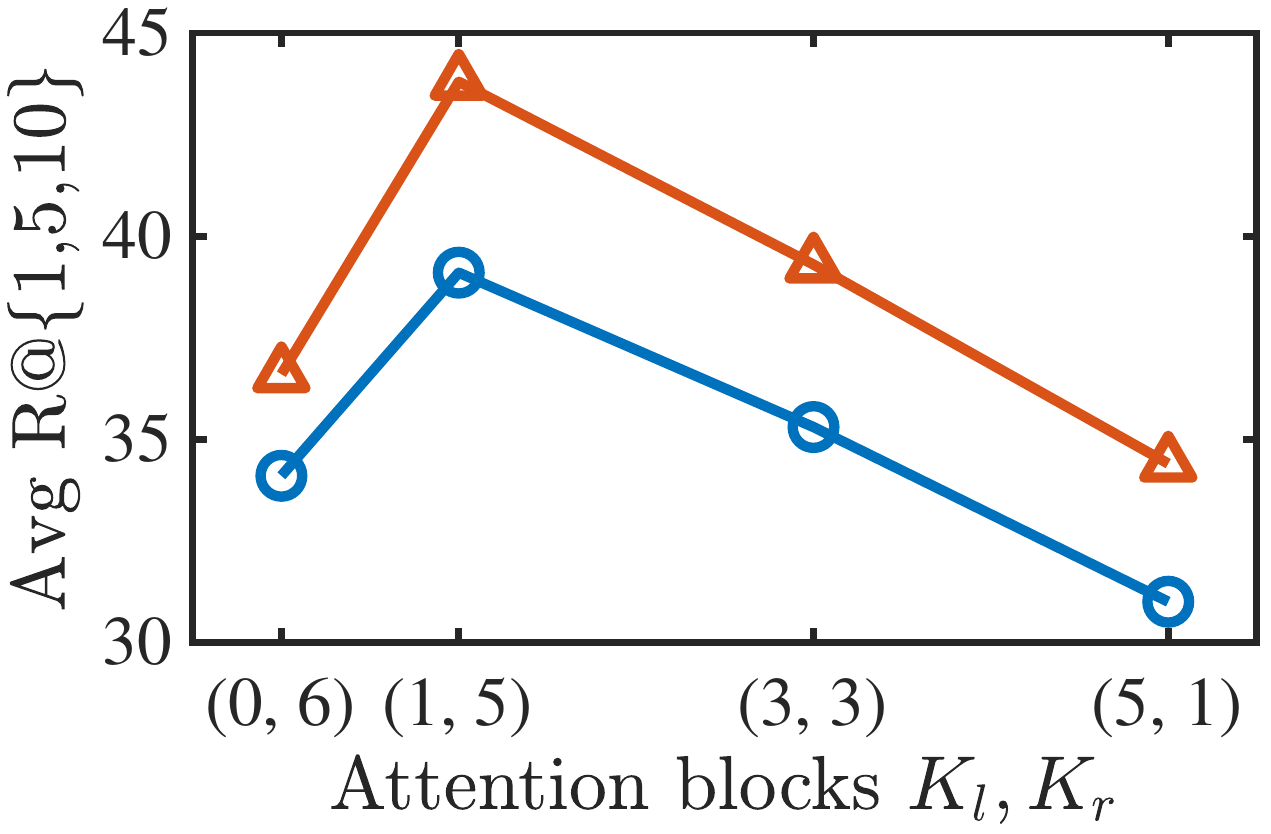}
    \caption{Number of local / random blocks.}
    \label{fig:sparse-attn-blocks}
    \end{subfigure}
    \begin{subfigure}[b]{.49\linewidth}
    \centering
    \includegraphics[height=0.6\linewidth]{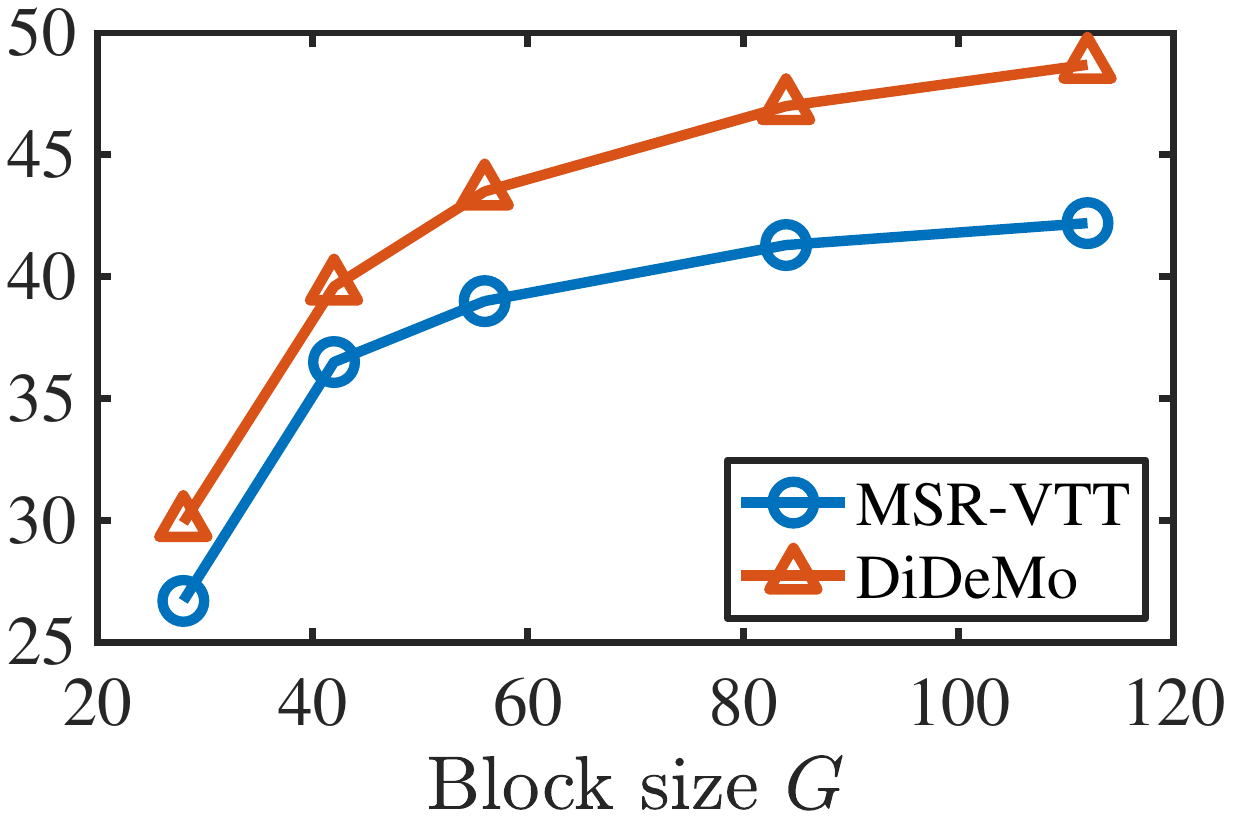}
    \caption{Attention block size.}
    \label{fig:sparse-attn-block-size}
    \end{subfigure}
    
    \caption{\textbf{Ablation on Edge Sparsity.} We evaluate dense model using different local/random blocks $(K_l, K_r)$ and block size $G$.}
    \label{fig:sparse-block-attn}
\end{figure}

%% file: assets/table/sparse-model-ablation.tex
\begin{table}[t]
    \centering
    \resizebox{\linewidth}{!}{
    
\begin{tabular}{@{}clcccccc@{}}
\toprule
 &  & \textbf{FLOPs} & \textbf{Mem.} & \multicolumn{4}{c}{\textbf{DiDeMo}} \\
\multirow{-2}{*}{$T$} & \multirow{-2}{*}{\textbf{Spars.}} & (G) & (GB) & R1 & R5 & R10 & \textbf{Mean} \\ \midrule
 & Dense & 139.9 & 0.96 & 28.8 & 53.1 & 63.0 & 48.3 \\
 & Edge & 135.9 & 0.80 & 28.3 & 51.9 & 61.4 & 47.2 \\
 & Node & 95.8 & 0.61 & \textbf{29.9} & \textbf{54.8} & \textbf{63.9} & \textbf{49.6} \\
\multirow{-4}{*}{4} & Hybrid & \textbf{93.9} & \textbf{0.54} & 29.3 & 53.7 & 63.2 & 48.8 \\ \midrule
 & Dense & 291.3 & 3.14 & 29.6 & 54.1 & 64.1 & 49.3 \\
 & Edge & 271.8 & 1.57 & 29.8 & 54.9 & 65.4 & 50.1 \\
 & Node & 197.6 & 1.77 & 30.4 & 55.7 & 66.0 & 50.7 \\
\multirow{-4}{*}{8} & Hybrid & \textbf{166.4} & \textbf{0.92} & \textbf{31.0} & \textbf{57.2} & \textbf{66.3} & \textbf{51.5} \\ \midrule
 & Dense & 627.8 & 10.66 & \multicolumn{4}{c}{\cellcolor[HTML]{C0C0C0}Untrainable} \\
 & Edge & 543.6 & 3.02 & \textbf{31.6} & 55.1 & 64.6 & 50.5 \\
 & Node & 370.9 & 4.39 & \multicolumn{4}{c}{\cellcolor[HTML]{C0C0C0}Untrainable} \\
\multirow{-4}{*}{16} & Hybrid & \textbf{296.2} & \textbf{1.57} & 31.4 & \textbf{57.3} & \textbf{67.8} & \textbf{52.2} \\ \bottomrule
\end{tabular}

    }
    \caption{\textbf{Ablation on Training Sparse Models}. We compare the zero-shot performance, inference GFLOPs, and training memory (per sample) of sparse models to the dense attention baseline. }
    \label{tab:sparse-model-ablation}
\end{table}

%% file: assets/figure/node-keep-rate.tex
\begin{figure}[t]
    \centering
    
    \begin{subfigure}[b]{.49\linewidth}
    \centering
    \includegraphics[height=0.6\linewidth]{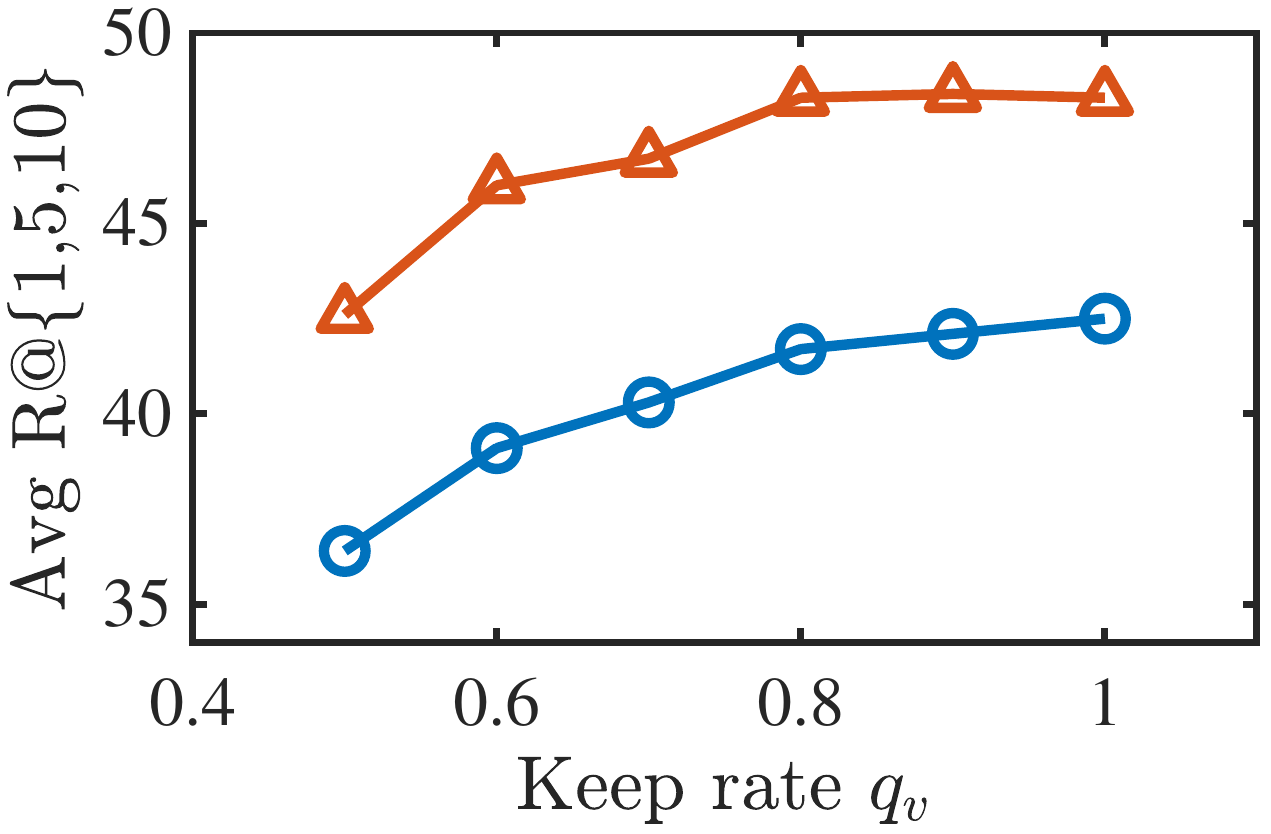}
    \caption{Visual sparsification.}
    \label{fig:visual-keep-rate}
    \end{subfigure}
    \begin{subfigure}[b]{.49\linewidth}
    \centering
    \includegraphics[height=0.6\linewidth]{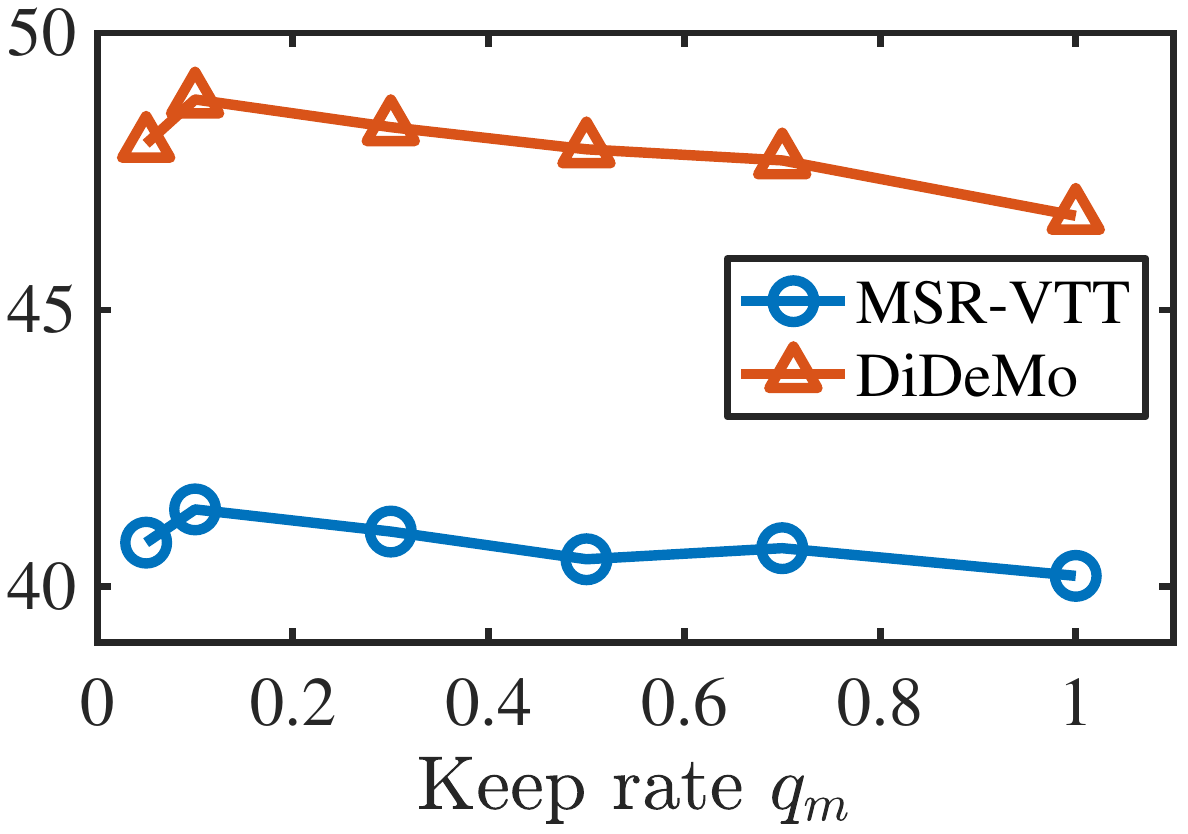}
    \caption{Multimodal sparsification.}
    \label{fig:multimodal-keep-rate}
    \end{subfigure}
    
    \caption{\textbf{Ablation on Node Sparsity.} We evaluate the pre-trained dense model using different keep rates $q$ at test time.}
    \label{fig:node-keep-rate}
\end{figure}

%% file: assets/table/sparse-expansion.tex
\begin{table}[t]
    \centering
    \resizebox{\linewidth}{!}{
\begin{tabular}{@{}crcccc@{}}
\toprule
\textbf{Frames} & \multicolumn{1}{c}{\textbf{Sparsity}} & \multicolumn{4}{c}{\textbf{DiDeMo}} \\
$T$ & \multicolumn{1}{c}{$T_S$} & R1 & R5 & R10 & \textbf{Mean} \\ \midrule
\multirow{2}{*}{$4$} & $4_{.80}$ & 28.0 & 50.7 & 59.2 & 46.0 \\
 & $1_{0} \to 4_{.80}$ & \textbf{29.3} & \textbf{53.7} & \textbf{63.2} & \textbf{48.8} \\ \midrule
\multirow{2}{*}{$8$} & $8_{.91}$ & 27.3 & 51.4 & 63.8 & 47.5 \\
 & $4_{.80} \to 8_{.91}$ & \textbf{31.0} & \textbf{57.2} & \textbf{66.3} & \textbf{51.5} \\ \midrule
\multirow{2}{*}{$16$} & $16_{.96}$ & 27.5 & 52.4 & 63.2 & 47.7 \\
 & $4_{.80} \to 8_{.91} \to 16_{.96}$ & \textbf{31.4} & \textbf{57.3} & \textbf{67.8} & \textbf{52.2} \\ \bottomrule
\end{tabular}
    }
    \caption{\textbf{Ablation on Temporal Sparse Expansion}. Sparsity $T_S$ indicates training on $T$ frames while removing $S \times 100\%$ attention edges of dense transformer.}
    \label{tab:sparse-expansion}
\end{table}

%% file: assets/table/zs-retrieval_v2.tex
\begin{table}[t]
    \centering
    \resizebox{\linewidth}{!}{
\begin{tabular}{@{}llllcccc@{}}
\toprule
\multicolumn{2}{l}{} &  &  & \multicolumn{2}{c}{\textbf{MSR-VTT}} & \multicolumn{2}{c}{\textbf{DiDeMo}} \\
\multicolumn{2}{l}{\multirow{-2}{*}{\textbf{Method}}} & \multirow{-2}{*}{\textbf{PT}} & \multirow{-2}{*}{$T$} & R1 & Mean & R1 & Mean \\ \midrule
\rowcolor[HTML]{EFEFEF} 
\multicolumn{2}{l}{\cellcolor[HTML]{EFEFEF}VideoCLIP~\cite{xu2021videoclip}} & 100M & --- & 10.4 & 20.9 & 16.6 & --- \\
\rowcolor[HTML]{EFEFEF} 
\multicolumn{2}{l}{\cellcolor[HTML]{EFEFEF}Frozen~\cite{bain2021frozen}} & \cellcolor[HTML]{EFEFEF}5M & 4 & 23.2 & 41.5 & 21.1 & 41.1 \\
\rowcolor[HTML]{EFEFEF} 
\multicolumn{2}{l}{\cellcolor[HTML]{EFEFEF}ALPRO~\cite{li2022align}} & \cellcolor[HTML]{EFEFEF}5M & 8 & 24.1 & 41.4 & 23.8 & 43.0 \\
\rowcolor[HTML]{EFEFEF} 
\multicolumn{2}{l}{\cellcolor[HTML]{EFEFEF}VIOLET~\cite{fu2021violet}} & 5M & 4 & 25.9 & 45.0 & 23.5 & 44.4 \\
\rowcolor[HTML]{EFEFEF} 
\multicolumn{2}{l}{\cellcolor[HTML]{EFEFEF}Singularity~\cite{lei2022revealing}} & \cellcolor[HTML]{EFEFEF}5M & 1 & \textit{28.4} & \textit{46.0} & \textit{36.9} & \textit{55.8} \\
\multicolumn{2}{l}{} &  & 1 & 21.1 & 38.7 & 23.3 & 40.8 \\
\multicolumn{2}{l}{} &  & 4 & 24.4 & 40.0 & 26.4 & 44.1 \\
\multicolumn{2}{l}{\multirow{-3}{*}{Singularity*}} & \multirow{-3}{*}{2M} & 8 & 24.3 & 41.0 & 25.8 & 45.5 \\ \midrule
 & Dense &  &  & \textbf{26.0} & 43.6 & 29.6 & 49.3 \\
\multirow{-2}{*}{\ours} & Hybrid & \multirow{-2}{*}{2M} & \multirow{-2}{*}{8} & 25.4 & \textbf{43.8} & \textbf{31.0} & \textbf{51.5} \\ \bottomrule
\end{tabular}
    }
    
    \caption{\textbf{Zero-shot Text-to-video Retrieval.} Results reported in prior works are marked in gray; * indicates our reproduced results.
    \textbf{PT} = \# video-text pairs for pre-training, $T$ = \# frames per clip. }
    \label{tab:zs-retrieval}
\end{table}

%% file: assets/figure/qualitative.tex
\begin{figure*}[t]
    \centering
    \includegraphics[width=0.9\linewidth]{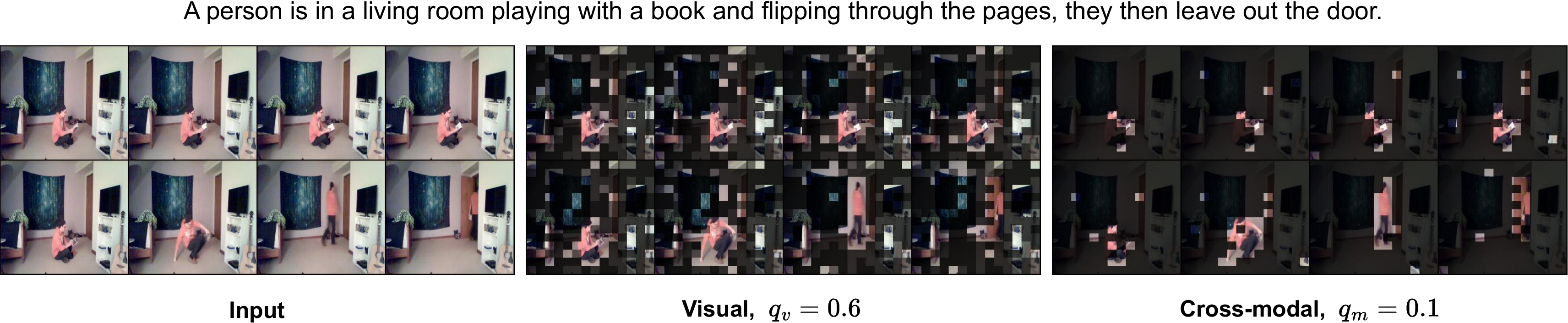}
    \caption{\textbf{Qualitative Results.} \ours\ isolates informative regions from background patches to facilitate efficient temporal reasoning. 
    }
    \label{fig:qualitative}
\end{figure*}

%% file: assets/table/ft-retrieval_v2.tex
\begin{table}[t]
    \centering

\resizebox{\linewidth}{!}{
\begin{tabular}{@{}llllcccc@{}}
\toprule
\multicolumn{2}{l}{} &  &  & \multicolumn{2}{c}{\textbf{Charades}} & \multicolumn{2}{c}{\textbf{SSv2}} \\
\multicolumn{2}{l}{\multirow{-2}{*}{\textbf{Method}}} & \multirow{-2}{*}{\textbf{PT}} & \multirow{-2}{*}{$T$} & R1 & Mean & R1 & Mean \\ \midrule
\multicolumn{2}{l}{Frozen~\cite{bain2021frozen}} & 5M & 32 & 11.9 & 25.1 & --- & --- \\
\multicolumn{2}{l}{CLIP4Clip~\cite{luo2022clip4clip}} & 400M & 12 & 13.9 & 27.1 & 43.1 & 65.1 \\
\multicolumn{2}{l}{ECLIPSE~\cite{lin2022eclipse}} & 400M & 32 & 15.7 & 30.3 & --- & --- \\
\multicolumn{2}{l}{MKTVR$^\dagger$~\cite{madasu2022improving}} & 400M & 42 & 16.6 & 34.7 & --- & \textbf{---} \\
\multicolumn{2}{l}{} &  & 1 & --- & --- & 36.4 & 58.9 \\
\multicolumn{2}{l}{\multirow{-2}{*}{Singularity~\cite{lei2022revealing}}} & \multirow{-2}{*}{5M} & 4 & --- & --- & 44.1 & 66.6 \\ \midrule
 & Dense &  &  & 16.0 & 32.7 & 43.6 & 66.1 \\
\multirow{-2}{*}{\ours} & Hybrid & \multirow{-2}{*}{2M} & \multirow{-2}{*}{8} & \textbf{17.7} & \textbf{35.7} & \textbf{49.8} & \textbf{69.3} \\ \bottomrule
\end{tabular}
}
    
    \caption{\textbf{Text-to-video Retrieval with Fine-tuning.}}
    \label{tab:ft-retrieval}
\end{table}

%% file: assets/table/video-qa_v1.tex
\begin{table}[t]
    \centering

\resizebox{\linewidth}{!}{
\begin{tabular}{@{}llllccc@{}}
\toprule
\multicolumn{2}{l}{\textbf{Method}} & \textbf{PT} & $T$ & \textbf{MSRVTT} & \textbf{ANet} & \textbf{AGQA} \\ \midrule
\multicolumn{2}{l}{HME~\cite{fan2019heterogeneous}} &  & 20 & 33.0 & --- & 47.7 \\
\multicolumn{2}{l}{HCRN~\cite{le2020hierarchical}} & \multirow{-2}{*}{---} & 128 & 35.5 & --- & 47.4 \\
\multicolumn{2}{l}{ClipBERT~\cite{lei2021less}} & 0.2M & 16 & 37.4 & --- & --- \\
\multicolumn{2}{l}{ALPRO~\cite{li2022align}} & 5M & 16 & 42.1 & --- & --- \\
\multicolumn{2}{l}{Just Ask~\cite{yang2021just}} & 69M & 640 & 41.5 & 38.9 & --- \\
\multicolumn{2}{l}{MERLOT~\cite{zellers2021merlot}} & 180M & 5 & 43.1 & 41.4 & --- \\
\multicolumn{2}{l}{VIOLET~\cite{fu2021violet}} & 185M & 4 & \textbf{43.9} & --- & 49.2 \\
\multicolumn{2}{l}{Singularity~\cite{lei2022revealing}} & 5M & 1 & 42.7 & 41.8 & --- \\ \midrule
 & Dense &  &  & 42.7 & 42.3 & 50.2 \\
\multirow{-2}{*}{\ours} & Hybrid & \multirow{-2}{*}{2M} & \multirow{-2}{*}{8} & 43.0 & \textbf{43.2} & \textbf{52.7} \\ \bottomrule
\end{tabular}
}
    
    \caption{\textbf{Video Question Answering.}}
    \label{tab:video-qa}
\end{table}

%% file: main/6_conclusion.tex
\vspace{-2mm}
\section{Conclusion}
\label{sec:6_conclusion}

This work introduced \ours{}, a sparse video-text transformer for efficient reasoning over a long temporal context. By interpreting visual transformers as graph networks, we proposed to optimize their \emph{edge} and \emph{node} sparsity, using a combination of sparse block attention, visual token pruning and text-guided token selection. We further introduced a temporal expansion strategy for training \ours{}, which aims to gradually increase model sparsity with clip length. \ours{} showed strong performance and efficiency compared to dense transformers, with a larger gap when frame number increases. On video retrieval and question answering benchmarks, \ours{} achieved state-of-the-art results using only video data, without extra image-text pretraining.

%% file: supp/1_implementation.tex
\section{Implementation Details}
\label{supp:implementation}

\subsection{Model Architecture}

\paragraph{Sparse configurations.}
The sparsity of \ours\ is controlled by the following hyperparameters: 
\begin{itemize}
    \item Visual token keep rate $q_v^{(l)}$ and multimodal token keep rate $q_m^{(l)}$ per layer $l$ for \emph{node} sparsity;
    \item Local attention blocks $K_l$, random attention blocks $K_r$ and block size $G$ shared across layers for \emph{edge} sparsity.
\end{itemize}

\cref{tab:model-details} lists the configurations for each stage of pre-training and the corresponding sparsity $s$, computed as the percent of reduction in edges of sparsified attention graph $\gG$ from that of a dense transformer. For the $l^\textrm{th}$ layer of visual encoder $f_v$, the number of edges is given by
\begin{equation}
    \abs{\gE_v^{(l)}} = N_v^{(l)} (K_l + K_r) G
\end{equation}
where input length $N_v^{(l)} = \ceil{q_v^{(l-1)} N_v^{(l-1)}}$. For multimodal layers $f_m$, the edge count is
\begin{equation}
    \abs{\gE_m^{(l)}} = N_m^{(l)} N_t
\end{equation}
where $N_t$ denotes text length and $N_m^{(l)} = \ceil{q_m^{(l-1)} N_m^{(l-1)}}$.
Therefore an \ours\ model with $L_v = 12$ visual layers and $L_m = 3$ multimodal layers has overall edge sparsity
\begin{equation}
    S(q_v, q_m, K_l, K_r) = 1 - \frac{\sum_{l=1}^{L_v} \abs{\gE_v^{(l)}} + \sum_{l=1}^{L_m} \abs{\gE_m^{(l)}}}{L_v N_v^2 + L_m N_t N_v}
    \label{eqn:sparsity}
\end{equation}

\input{assets/table/supp_arch.tex}

\paragraph{Temporal expansion.}
Transformer architectures do not require fixed input lengths as its operations are either pointwise (e.g. FFN) or permutation equivariant (e.g. MHSA). This makes the temporal expansion (Sect.~4) of input clips a mostly trivial process, except for the position embeddings, which does depend on spatiotemporal dimensions of inputs. Following prior work on training video transformers with image models, we \emph{inflate} the 2D positional embedding
\begin{equation}
    \rmP = [\rvp_\cls, \rvp_{1, 1}, \dots, \rvp_{H, W}] \in \R^{(HW+1) \times d}
\end{equation}
into a 3D embedding tensor
\begin{equation}
    \rmP' = [\rvp_\cls, \rvp'_{1, 1, 1}, \dots, \rvp'_{T, H, W}] \in \R^{(THW+1) \times d}
\end{equation}
for inputs of $T$ frames, by duplicating the local embeddings $\rvp_{hw}$ along the temporal dimension:
\begin{equation}
    \rvp'_{t, h, w} = \rvp_{h, w}, \qquad \forall t, h, w
\end{equation}

Likewise, expansion of clip length from $T_1$ to $T_2$ can be performed by temporally resizing the positional embedding, e.g. through nearest neighbors interpolation:
\begin{equation}
    \rvp'_{t, h, w} = \rvp_{\floor{t \cdot \frac{T_1}{T_2} + \frac{1}{2}}, h, w}, \qquad \forall t, h, w
\end{equation}

The BEiT backbone of visual encoder uses relative position bias~\cite{raffel2020exploring} in every self-attention layer, which encodes a scalar added to each entry of the similarity matrix depending on the relative position between query and key patches:
\begin{gather}
    \gA(\rmQ, \rmK, \rmV) = \sigma(\rmQ \rmK^T + \rmB) \rmV, \\[1mm]
    \rmB_{(h, w), (h', w')} = \rmR_{h'-h, w'-w}
\end{gather}
where $\rmR \in \R^{(2H-1) \times (2W-1)}$ are learnable parameters. When expanding the input to multi-frame clips, we again inflate the relative position bias to the temporal dimension:
\begin{gather}
    \rmB'_{(t, h, w), (t', h', w')} = \rmR'_{t'-t, h'-h, w'-w}, \\[1mm]
    \rmR' \in \R^{(2T-1) \times (2H-1) \times (2W-1)}
\end{gather}
$\rmR'$ is initialized by interpolating $\rmR$ temporally, identical to the procedure for absolute positional embedding $\rmP'$.

\subsection{Datasets}

\input{assets/table/supp_datasets.tex}

\paragraph{Pre-training.}
\ours\ is pre-trained on WebVid-2M~\cite{bain2021frozen} with 2.5 million video-text pairs scraped from the Internet. While alternative datasets exist for video-language pre-training such as HowTo100M~\cite{miech2019howto100m} and YT-Temporal~\cite{zellers2021merlot}, we choose WebVid as it has higher caption quality, covers a wide range of scenes, and can be trained with a reasonable amount of resource.

\paragraph{Text-to-video retrieval.}
We evaluate text-to-video retrieval on 4 datasets: MSR-VTT~\cite{xu2016msr}, DiDeMo~\cite{anne2017localizing}, Charades~\cite{sigurdsson2016hollywood} and Something-Something v2~\cite{goyal2017something}. MSR-VTT and DiDeMo are video-text datasets commonly used in prior work; Charades and SSv2 were initially collected for video action recognition, with an emphasis on human-object interactions and temporal modeling, but also includes text descriptions for each video clip. 

\paragraph{Video question answering.}
Video question answering is evaluated on MSRVTT-QA~\cite{xu2017video}, ActivityNet-QA~\cite{yu2019activitynet} and AGQA 2.0~\cite{grunde2022agqa}, annotated on top of the videos from MSR-VTT~\cite{xu2016msr}, ActivityNet~\cite{caba2015activitynet} and Charades~\cite{sigurdsson2016hollywood} respectively. MSRVTT-QA consists of mostly descriptive questions which can be solved without intricate temporal reasoning. ActivityNet-QA focuses on human actions and spatiotemporal relation between objects, posing a greater challenge beyond frame-based reasoning. AGQA contains difficult questions involving the composition of actions, testing the generalization capacity of video-text models.

\cref{tab:datasets} summarizes the statistics of all aforementioned datasets.

\subsection{Training Details}

\input{assets/table/supp_hyperparams.tex}
\input{assets/table/supp_zs-retrieval.tex}

\paragraph{Pre-training tasks.}
\ours\ is pre-trained on three losses following prior art in VLP~\cite{chen2020uniter,li2020hero,li2021align,fu2021violet,lei2022revealing}.
\begin{itemize}
    \item Video-text contrastive (VTC) applies InfoNCE loss between the video embeddings $\rmZ_v$ and text embeddings $\rmZ_t$ extracted at \texttt{[cls]} locations of their respective encoder $f_v$ and $f_t$:\footnote{Linear projection on top of $\rvz_v, \rvz_t$ omitted.}
    \begin{gather}
        \gL_\textrm{VTC} = \ell_c(\rmZ_v, \rmZ_t) + \ell_c(\rmZ_t, \rmZ_v), \\
        \ell_c(\rmX, \rmY) = -\sum_{i=1}^B \log \frac{e^{\inner{\rvx_i, \rvy_i} / \tau}}{\sum_{j=1}^B e^{\inner{\rvx_i, \rvy_j} / \tau}}
    \end{gather}

    \item Video-text matching (VTM) learns a binary classifier on top of the \texttt{[cls]} output of multimodal encoder $f_m$ to discriminate between paired and misaligned video-text pair, optimized by binary cross entropy:
    \begin{align}
        \gL_\textrm{VTM} = -\sum_{i=1}^B \Big(&\log(f_m(\rvz_{v, i}, \rvz_{t, i})) \nonumber \\
        & + \log(1 - f_m(\rvz_{v, i}, \rvz_{t, i'}))\Big)
    \end{align}
    where $i' \neq i$ is a randomly selected negative sample.

    \item Masked language modeling (MLM) requires the multimodal encoder $f_m$ to predict randomly masked out text tokens conditioned on the rest of text and video sequence, through a cross-entropy loss:
    \begin{gather}
        \gL_\textrm{MLM} = -\sum_{i=1}^B \sum_{j \in \gJ} [\rvx_t]_{i, j}^T \log \rvy_{i, j}
    \end{gather}
    where $[\rvx_t]_{i, j}$ is a one-hot vector denoting the word at location $j$ of example $i$, $\rvy_{i, j}$ is the classifier output predicting the word at the same location, and $\gJ$ is the set of masked indices.
\end{itemize}

We use equal weights for all three losses.

\paragraph{Downstream tasks.}
We follow the downstream evaluation setup of Singularity~\cite{lei2022revealing} for the most part.
Text-to-video retrieval is performed by ranking all candidate videos $\rvx_v$ of the test set by their matching scores to text query $\rvx_t$. For video QA, a transformer decoder is applied on top of multimodal encoder $f_m$ to generate the answer.

\paragraph{Training hyper-parameters.}
We use a sparse frame sampling strategy following~\cite{bain2021frozen,fu2021violet,lei2022revealing}, splitting input videos into $T$ chunks and randomly selecting one frame from each during training.
Video frames are preprocessed with random resized cropping into spatial resolution of $224 \times 224$, resulting in $14 \times 14$ spatial patches.
All models are optimized using AdamW~\cite{loshchilov2017decoupled} ($\beta_1 = 0.9$, $\beta_2 = 0.999$) with a cosine learning rate schedule and warm-up training. 
We use 10 epochs for pre-training and 15 for fine-tuning on all datasets other than AGQA, which uses 1 epoch due to its large size. Batch size $B$ and learning rate $\eta$ are adjusted depending on memory costs of sparse models. 
\cref{tab:hyperparams} summarizes the hyperparameters used for each task and model variant.

%% file: assets/table/supp_arch.tex
\begin{table}[t]
    \centering
    \resizebox{\linewidth}{!}{
\begin{tabular}{@{}ccccc@{}}
\toprule
\textbf{Frames} & \textbf{Attn. blocks} & \textbf{Keep rate} & \textbf{Edges (M)} & \textbf{Sparsity} \\
$T$ & $K_l, K_r, G$ & $q_v, q_m$ & $\abs{\gE}$ & $S$ \\ \midrule
4 & \multirow{3}{*}{(1, 3, 56)} & (0.7, 0.1) & 1.48 & 0.80 \\
8 &  & (0.6, 0.1) & 2.60 & 0.91 \\
16 &  & (0.5, 0.1) & 4.61 & 0.96 \\ \bottomrule
\end{tabular}
    }
    \caption{\textbf{\ours\ Configurations.} We report hyperparameters controlling the edge and node sparsity for different clip lengths $T$, as well as the overall sparsity as computed by \cref{eqn:sparsity}.}
    \label{tab:model-details}
\end{table}

%% file: assets/table/supp_datasets.tex
\begin{table}[t]
    \centering
    \resizebox{\linewidth}{!}{
    \begin{tabular}{@{}lcrr@{}}
    \toprule
    \textbf{Dataset} & \textbf{Avg. Dur.} & \textbf{\# Videos} & \textbf{\# Sent. / Q.} \\ \midrule
    \multicolumn{4}{c}{\textit{Video-Text Pre-Training}}  \\ \midrule
    WebVid-2M~\cite{bain2021frozen}         & 18s       & 2.5M      & 2.5M      \\ \midrule
    \multicolumn{4}{c}{\textit{Text-to-Video Retrieval}}  \\ \midrule
    MSR-VTT~\cite{xu2016msr}                & 15s       & 10K       & 200K      \\
    DiDeMo~\cite{anne2017localizing}        & 28s       & 10K       & 40K       \\
    Charades~\cite{sigurdsson2016hollywood} & 30s       & 10K       & 16K       \\
    SSv2-Label~\cite{lei2022revealing}      & 4s        & 171K      & 112K      \\ \midrule
    \multicolumn{4}{c}{\textit{Video Question Answering}} \\ \midrule
    MSRVTT-QA~\cite{xu2017video}            & 15s       & 10K       & 244K      \\
    ActivityNet-QA~\cite{yu2019activitynet} & 180s      & 5.8K      & 58K       \\
    AGQA 2.0~\cite{grunde2022agqa}          & 30s       & 10K       & 2.27M     \\ \bottomrule
    \end{tabular}
    }
    \caption{\textbf{Pre-Training and Downstream Datasets.} }
    \label{tab:datasets}
\end{table}

%% file: assets/table/supp_hyperparams.tex
\begin{table*}[t]
    \centering
    \resizebox{0.9\linewidth}{!}{
\begin{tabular}{@{}l
>{\columncolor[HTML]{EFEFEF}}c 
>{\columncolor[HTML]{EFEFEF}}c 
>{\columncolor[HTML]{EFEFEF}}c ccc
>{\columncolor[HTML]{EFEFEF}}c @{}}
\toprule
\textbf{Task} & \multicolumn{3}{c}{\cellcolor[HTML]{EFEFEF}\textbf{Pre-training}} & \multicolumn{3}{c}{\textbf{Video-text Retrieval}} & \textbf{Video QA} \\
\textbf{Frames $T$} & 4 & 8 & 16 & 4 & 8 & 16 & 8 \\ \midrule
\textbf{Epochs} & \multicolumn{3}{c}{\cellcolor[HTML]{EFEFEF}10} & \multicolumn{3}{c}{15} & 5 (1 for AGQA) \\
\textbf{Warm-up} & \multicolumn{3}{c}{\cellcolor[HTML]{EFEFEF}1} & \multicolumn{3}{c}{0} & 0 \\
\textbf{Batch size} & 512 & 336 & 192 & 64 & 48 & 32 & 128 \\
\textbf{Learning rate} & $3 \times 10^{-5}$ & $1 \times 10^{-5}$ & $5 \times 10^{-6}$ & \multicolumn{3}{c}{$1 \times 10^{-5}$} & $5 \times 10^{-5}$ \\
\textbf{Weight decay} & \multicolumn{3}{c}{\cellcolor[HTML]{EFEFEF}0.02} & \multicolumn{3}{c}{0.02} & 0.02 \\ \midrule
\textbf{Text length} & \multicolumn{3}{c}{\cellcolor[HTML]{EFEFEF}32} & \multicolumn{3}{c}{32 (64 for DiDeMo)} & 25 (Q), 5 (A) \\
\textbf{Attn. blocks $(K_l, K_r, G)$} & \multicolumn{3}{c}{\cellcolor[HTML]{EFEFEF}$(1, 3, 56)$} & \multicolumn{3}{c}{$(1, 3, 56)$} & $(1, 3, 56)$ \\
\textbf{Keep rate $(q_v, q_m)$} & $(0.7, 0.1)$ & $(0.6, 0.1)$ & $(0.5, 0.1)$ & $(0.7, 0.1)$ & $(0.6, 0.1)$ & $(0.5, 0.1)$ & $(0.6, 0.5)$ \\ \bottomrule
\end{tabular}
    }
    \caption{\textbf{Training Hyperparameters.} }
    \label{tab:hyperparams}
\end{table*}

%% file: assets/table/supp_zs-retrieval.tex
\begin{table*}[t]
    \centering
    \resizebox{0.9\linewidth}{!}{
\begin{tabular}{@{}llllcccccccc@{}}
\toprule
 &  &  &  & \multicolumn{4}{c}{\textbf{MSR-VTT}} & \multicolumn{4}{c}{\textbf{DiDeMo}} \\
\multirow{-2}{*}{\textbf{Method}} & \multirow{-2}{*}{\textbf{PT}} & \multirow{-2}{*}{\textbf{Frames}} & \multirow{-2}{*}{\textbf{Sparsity}} & R1 & R5 & R10 & \textbf{Mean} & R1 & R5 & R10 & \textbf{Mean} \\ \midrule
\rowcolor[HTML]{EFEFEF} 
VideoCLIP~\cite{xu2021videoclip} & 100M & --- & \cellcolor[HTML]{EFEFEF} & 10.4 & 22.2 & 30.0 & \textbf{20.9} & 16.6 & 46.9 & --- & \textbf{---} \\
\rowcolor[HTML]{EFEFEF} 
Frozen~\cite{bain2021frozen} & 5M & 4 & \cellcolor[HTML]{EFEFEF} & 23.2 & 44.6 & 56.6 & \textbf{41.5} & 21.1 & 46.0 & 56.2 & \textbf{41.1} \\
\rowcolor[HTML]{EFEFEF} 
ALPRO~\cite{li2022align} & 5M & 8 & \cellcolor[HTML]{EFEFEF} & 24.1 & 44.7 & 55.4 & \textbf{41.4} & 23.8 & 47.3 & 57.9 & \textbf{43.0} \\
\rowcolor[HTML]{EFEFEF} 
VIOLET~\cite{fu2021violet} & 5M & 4 & \cellcolor[HTML]{EFEFEF} & \multicolumn{1}{l}{\cellcolor[HTML]{EFEFEF}25.9} & \multicolumn{1}{l}{\cellcolor[HTML]{EFEFEF}49.5} & \multicolumn{1}{l}{\cellcolor[HTML]{EFEFEF}59.7} & \textbf{45.0} & \multicolumn{1}{l}{\cellcolor[HTML]{EFEFEF}23.5} & \multicolumn{1}{l}{\cellcolor[HTML]{EFEFEF}49.8} & \multicolumn{1}{l}{\cellcolor[HTML]{EFEFEF}59.8} & \textbf{44.4} \\
\rowcolor[HTML]{EFEFEF} 
Singularity~\cite{lei2022revealing} & 5M & 1 & \multirow{-5}{*}{\cellcolor[HTML]{EFEFEF}---} & 28.4 & 50.2 & 59.5 & \textbf{46.0} & 36.9 & 61.1 & 69.3 & \textbf{55.8} \\
 &  & 1 &  & 21.1 & 42.1 & 53.0 & \textbf{38.7} & 23.3 & 45.4 & 53.7 & \textbf{40.8} \\
 &  & 4 &  & 24.4 & 43.8 & 51.7 & \textbf{40.0} & 26.4 & 48.7 & 57.3 & \textbf{44.1} \\
\multirow{-3}{*}{Singularity*} & \multirow{-3}{*}{2M} & 8 & \multirow{-3}{*}{---} & 24.3 & 44.5 & 54.3 & \textbf{41.0} & 25.8 & 50.0 & 60.7 & \textbf{45.5} \\ \midrule
 &  &  & Dense & 26.0 & 47.7 & 57.1 & \textbf{43.6} & 29.6 & 54.1 & 64.1 & \textbf{49.3} \\
\multirow{-2}{*}{\ours} & \multirow{-2}{*}{2M} & \multirow{-2}{*}{8} & \cellcolor[HTML]{FFF2CC}Hybrid & \cellcolor[HTML]{FFF2CC}25.4 & \cellcolor[HTML]{FFF2CC}48.4 & \cellcolor[HTML]{FFF2CC}57.5 & \cellcolor[HTML]{FFF2CC}\textbf{43.8} & \cellcolor[HTML]{FFF2CC}31.0 & \cellcolor[HTML]{FFF2CC}57.2 & \cellcolor[HTML]{FFF2CC}66.3 & \cellcolor[HTML]{FFF2CC}\textbf{51.5} \\ \bottomrule
\end{tabular}
    }
    
    \caption{\textbf{Zero-shot Text-to-video Retrieval.} Results reported in prior works marked in gray; * indicates our reproduced results.}
    \label{tab:zs-retrieval-full}
\end{table*}

%% file: supp/2_results.tex
\section{Additional Results \& Analysis}
\label{supp:results}

\input{assets/table/supp_ft-retrieval.tex}
\input{assets/table/supp_frozen.tex}

\subsection{Retrieval Metrics}
We include full retrieval results with $\textrm{Recall}@\{1, 5, 10\}$ in \cref{tab:zs-retrieval-full} (zero-shot) and \cref{tab:ft-retrieval-full} (fine-tuned).

\input{assets/figure/supp_keep-rate.tex}

\subsection{Video-Text Backbone}
In addition to the Singularity baseline with BEiT-B backbone used in the main paper, we also evaluate \ours\ on a simpler structure from Frozen~\cite{bain2021frozen}. This is also a two-tower model with separate video and text encoders $f_v, f_t$, but unlike most vision-language transformers, does not contain a cross-modal encoder on top. Frozen is trained solely on the InfoNCE loss between video and text embeddings, and uses their cosine similarity to perform retrieval. While the cross-modal node sparsification does not apply to this framework, visual node sparsity and edge sparsity can still be applied to the visual encoder $f_v$ to enable temporal learning across frames.

The original Frozen model uses a divided space-time attention similar to TimeSformer~\cite{bertasius2021space}, where temporal attention is added to a pre-trained ViT and initialized as identity mapping. During early experiments, however, we find that the temporal module with zero-init fails to learn meaningful attention across frames, with query and key matrices stuck at zero weights. We opted to remove the temporal attention modules and make the spatial attention global instead (i.e. each token attends to every token from the video clip, instead of just those from the same frame). 

\cref{tab:zs-retrieval-frozen} shows the performance of \ours\ applied to the Frozen model. Similar to the results in the main paper, our dense spatiotemporal transformer with the above modifications outperformed the original implementation of~\cite{bain2021frozen}, despite being trained without image-text data (CC3M~\cite{sharma2018conceptual}). \ours\ with hybrid sparsity again outperforms the dense version while using less computation and training memory.

\subsection{Video Sparsity vs. Clip Length}

To demonstrate the claim that video sparsity increases with clip length, we evaluate dense models trained with clip length 4 and 8 under different levels of sparsity. As shown in \cref{fig:keep-rate-frames}, the 8-frame model is more robust to token pruning with lower keep rates. On DiDeMo, it outperforms 4-frame model by 4\% at $q_v = 0.5$, while the two models differ by under 2\% under dense evaluation. This reveals that longer clips contain greater level of redundancy, and should be modeled with higher sparsity (as done in this work).

\subsection{Chunking Strategy}
\input{assets/figure/supp_qualitative.tex}
\input{assets/table/supp_token-order.tex}

In edge sparsification, the flattened video sequence $\rvz_{1:N}$ is chunked into subsequences of length $G$. While this strategy is straightforward and common in language transformers~\cite{beltagy2020longformer,zaheer2020big}, it breaks the spatiotemporal continuity of video data. We investigate an alternative to na\"ive chunking, by reordering the input tokens using space-filling curves such as Morton~\cite{morton1966computer} and Hilbert~\cite{hilbert1891stetige} curves. This ensures that neighboring tokens in the flattened sequence are close to each other in the original multidimensional space, leading to more localized chunks.

However, early experiments showed no benefit of space-filling token order over na\"ive flattening, as shown in \cref{tab:token-order-ablation}. This is possibly because video encoders are initialized from image transformers, and block attention with reordering prevents video tokens from attending to other spatial locations from the same frame. We leave the study of an optimal chunking strategy for 3D inputs for future work.

\subsection{Temporal Probing}
\input{assets/table/temporal-probe.tex}

To measure the sensitivity of the learned video-text model to temporal cues, we perform an evaluation with shuffled input frames. \cref{tab:temporal-probe} shows a performance drop of \ours\ models on retrieval (SSv2) and video QA (ActivityNet) tasks, indicating that the video-text models have learned to reason about the temporal dynamics of video clips. The difference $\Delta$ between normal and shuffled inputs is more prominent on hybrid sparse models, possibly because they attend more to the foreground which contains more temporal variations.
Notably, this behavior does not hold for the Singularity model, whose performance is unaffected by frame order. This suggests that late temporal aggregation after spatial global pooing is insufficient to capture spatiotemporal relations across video frames.

\subsection{Qualitative Results}

\cref{fig:supp_qualitative,fig:supp_qualitative2} visualizes the node sparsification patterns generated by visual encoder $f_v$ and multimodal encoder $f_m$. While visual sparsification alone can significantly reduce the number of tokens during forward pass, we find that 
the cross-modal attention map aligns better with regions of interest in each clip, enabling greater node sparsity in video-text modeling.

%% file: assets/table/supp_ft-retrieval.tex
\begin{table*}[t]
    \centering

\resizebox{0.85\linewidth}{!}{
\begin{tabular}{@{}llllcccccccc@{}}
\toprule
 &  &  &  & \multicolumn{4}{c}{\textbf{Charades}} & \multicolumn{4}{c}{\textbf{SSv2-Label}} \\
\multirow{-2}{*}{\textbf{Method}} & \multirow{-2}{*}{\textbf{PT}} & \multirow{-2}{*}{\textbf{Frames}} & \multirow{-2}{*}{\textbf{Sparsity}} & R1 & R5 & R10 & \textbf{Mean} & R1 & R5 & R10 & \textbf{Mean} \\ \midrule
Frozen~\cite{bain2021frozen} & 5M & 32 &  & 11.9 & 28.3 & 35.1 & \textbf{25.1} & \multicolumn{4}{c}{---} \\
CLIP4Clip~\cite{luo2022clip4clip} & 400M & 12 &  & 13.9 & 30.4 & 37.1 & \textbf{27.1} & 43.1 & 71.4 & 80.7 & \textbf{65.1} \\
ECLIPSE~\cite{lin2022eclipse} & 400M & 32 &  & 15.7 & 32.9 & 42.4 & \textbf{30.3} & \multicolumn{4}{c}{---} \\
MKTVR$^\dagger$~\cite{madasu2022improving} & 400M & 42 &  & 16.6 & 37.5 & 50.0 & \textbf{34.7} & \multicolumn{4}{c}{---} \\
 &  & 1 &  & \multicolumn{4}{c}{---} & 36.4 & 64.9 & 75.4 & \textbf{58.9} \\
\multirow{-2}{*}{Singularity~\cite{lei2022revealing}} & \multirow{-2}{*}{5M} & 4 & \multirow{-6}{*}{---} & \multicolumn{4}{c}{---} & 44.1 & 73.5 & 82.2 & \textbf{66.6} \\ \midrule
 &  &  & Dense & 16.0 & 34.9 & 47.2 & \textbf{32.7} & 43.6 & 72.6 & 82.2 & \textbf{66.1} \\
\multirow{-2}{*}\ours & \multirow{-2}{*}{2M} & \multirow{-2}{*}{8} & \cellcolor[HTML]{FFF2CC}Hybrid & \cellcolor[HTML]{FFF2CC}17.7 & \cellcolor[HTML]{FFF2CC}39.5 & \cellcolor[HTML]{FFF2CC}49.8 & \cellcolor[HTML]{FFF2CC}\textbf{35.7} & \cellcolor[HTML]{FFF2CC}47.5 & \cellcolor[HTML]{FFF2CC}76.3 & \cellcolor[HTML]{FFF2CC}84.2 & \cellcolor[HTML]{FFF2CC}\textbf{69.3} \\ \bottomrule
\end{tabular}
}
    
    \caption{\textbf{Text-to-video Retrieval with Fine-tuning.} $^\dagger$ denotes concurrent work.}
    \label{tab:ft-retrieval-full}
\end{table*}

%% file: assets/table/supp_frozen.tex
\begin{table}[t]
    \centering
    \resizebox{\linewidth}{!}{
\begin{tabular}{@{}llllcccc@{}}
\toprule
\multicolumn{2}{l}{\multirow{2}{*}{\textbf{Method}}} & \multirow{2}{*}{\textbf{PT}} & \multirow{2}{*}{\textbf{Frames}} & \multicolumn{4}{c}{\textbf{DiDeMo}} \\
\multicolumn{2}{l}{} &  &  & R1 & R5 & R10 & \textbf{Mean} \\ \midrule
\multicolumn{2}{l}{Frozen~\cite{bain2021frozen}} & 5M & 4 & 21.1 & 46.0 & 56.2 & \textbf{41.1} \\ \midrule
\multirow{2}{*}{\ours} & Dense & \multirow{2}{*}{2M} & \multirow{2}{*}{8} & 21.9 & 45.6 & 56.6 & \textbf{41.4} \\
 & Hybrid &  &  & 22.9 & 47.7 & 58.1 & \textbf{42.9} \\ \bottomrule
\end{tabular}
    }
    
    \caption{\textbf{Zero-shot Retrieval with \ours\ on Frozen Baseline.}}
    \label{tab:zs-retrieval-frozen}
\end{table}

%% file: assets/figure/supp_keep-rate.tex
\begin{figure}[t]
    \centering
    
    \begin{subfigure}[b]{.49\linewidth}
    \centering
    \includegraphics[height=0.65\linewidth]{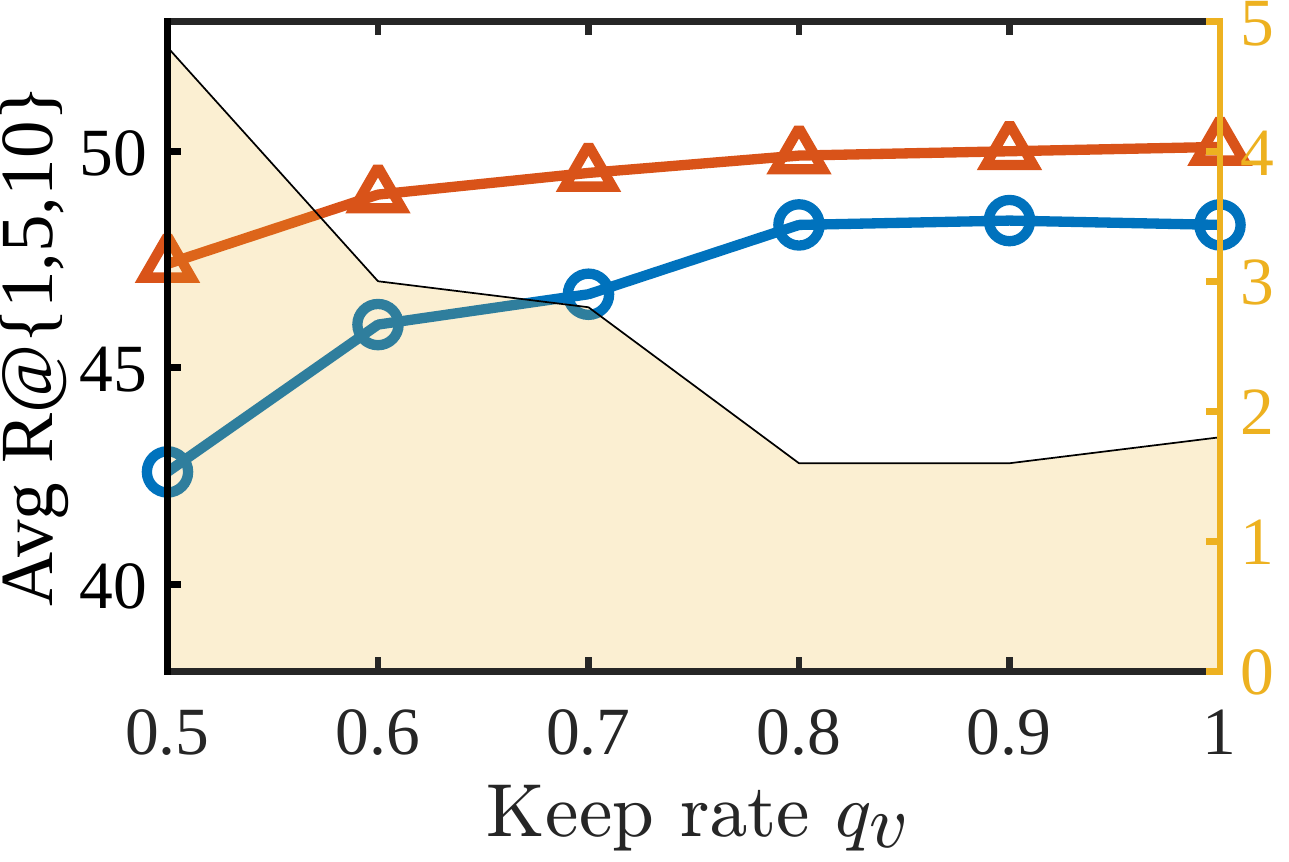}
    \caption{DiDeMo zero-shot.}
    \end{subfigure}
    \begin{subfigure}[b]{.49\linewidth}
    \centering
    \includegraphics[height=0.65\linewidth]{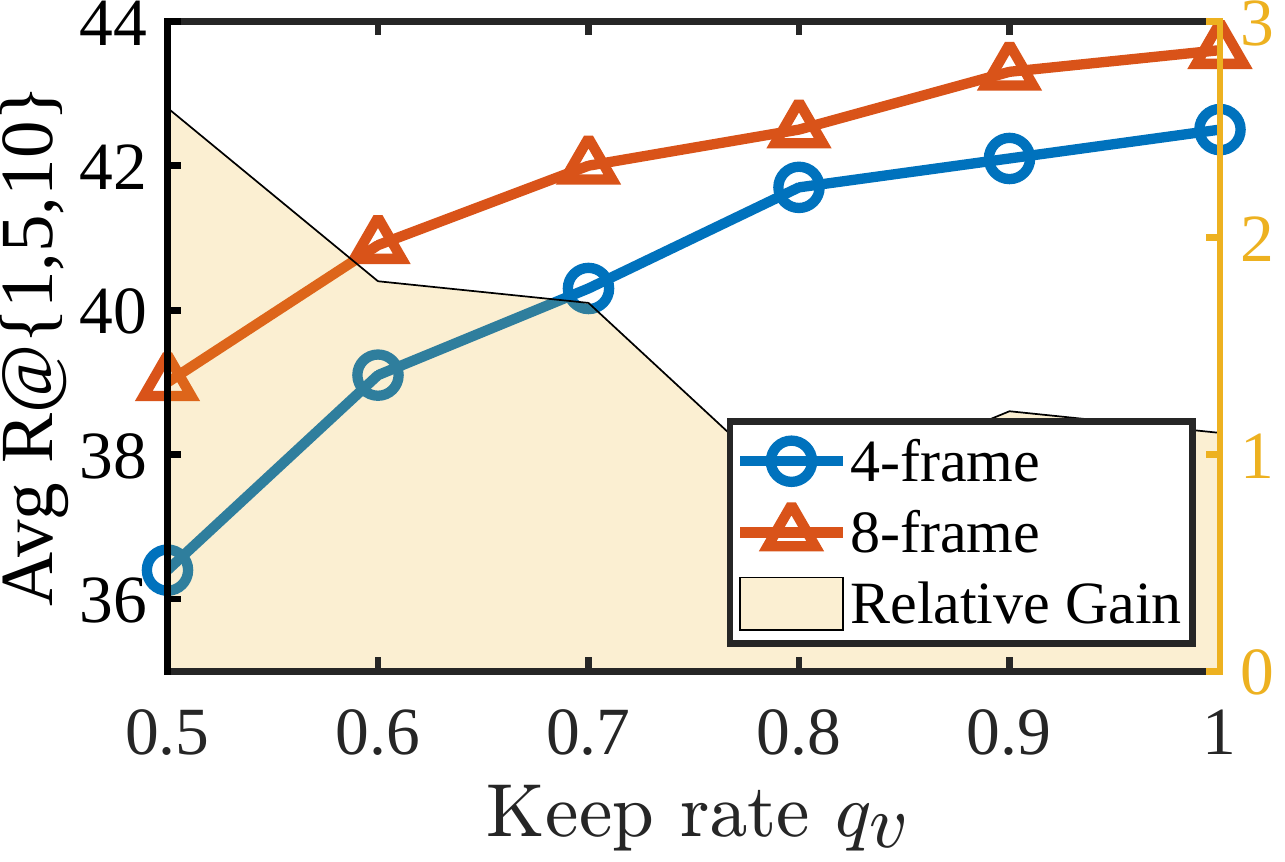}
    \caption{MSR-VTT zero-shot.}
    \end{subfigure}
    
    \caption{\textbf{Node Sparsity for 4- and 8-frame Models.} Model of longer clip length is more robust to node sparsification.}
    \label{fig:keep-rate-frames}
\end{figure}

%% file: assets/figure/supp_qualitative.tex
\begin{figure*}[t]
    \centering
    \includegraphics[width=0.95\linewidth]{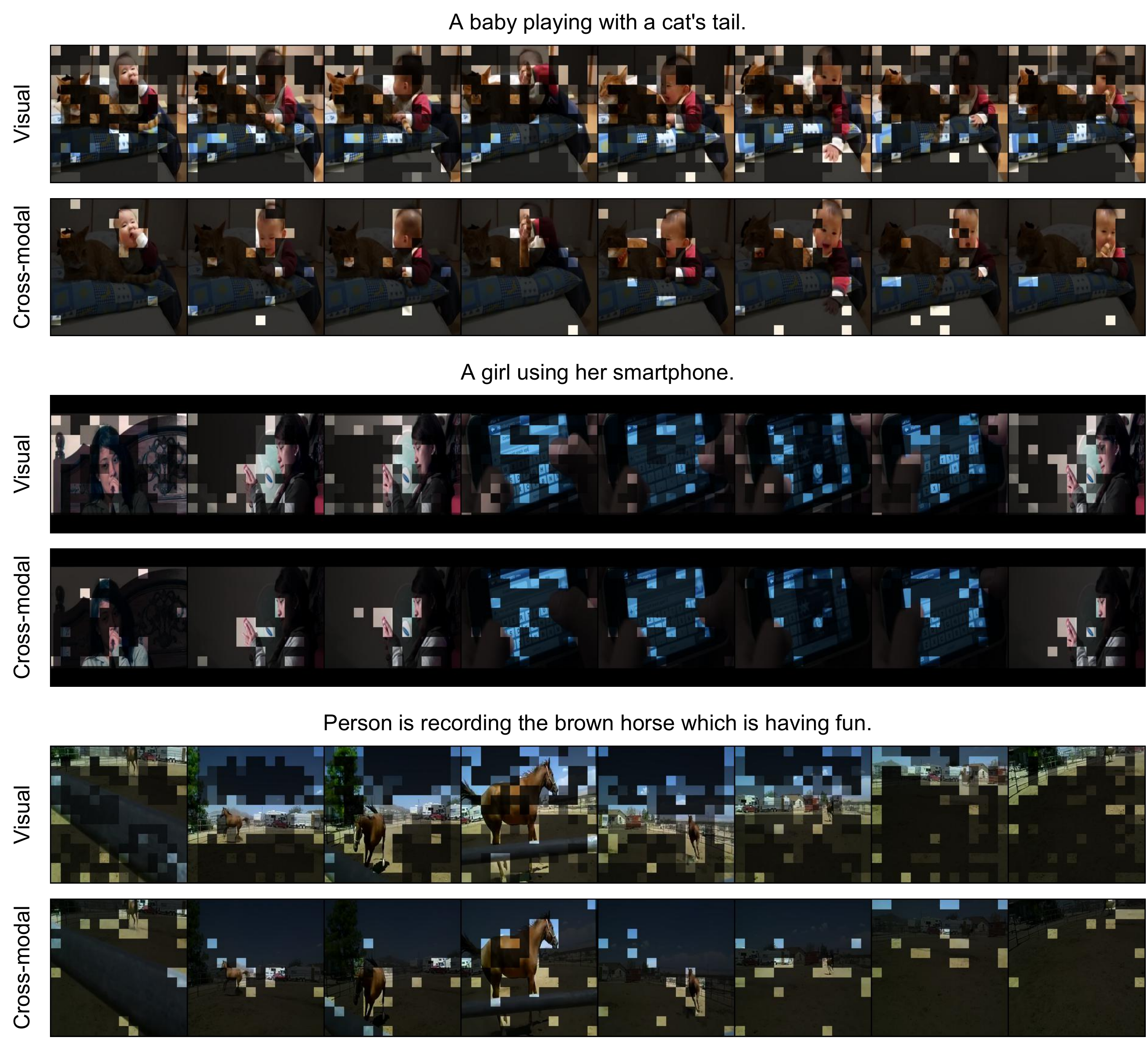}
    \caption{\textbf{Qualitative Results.} We visualize node sparsity patterns generated by visual ($q_v = 0.6$) and cross-modal encoder ($q_m = 0.1$).}
    \label{fig:supp_qualitative}
\end{figure*}

\begin{figure*}[t]
    \centering
    \includegraphics[width=0.95\linewidth]{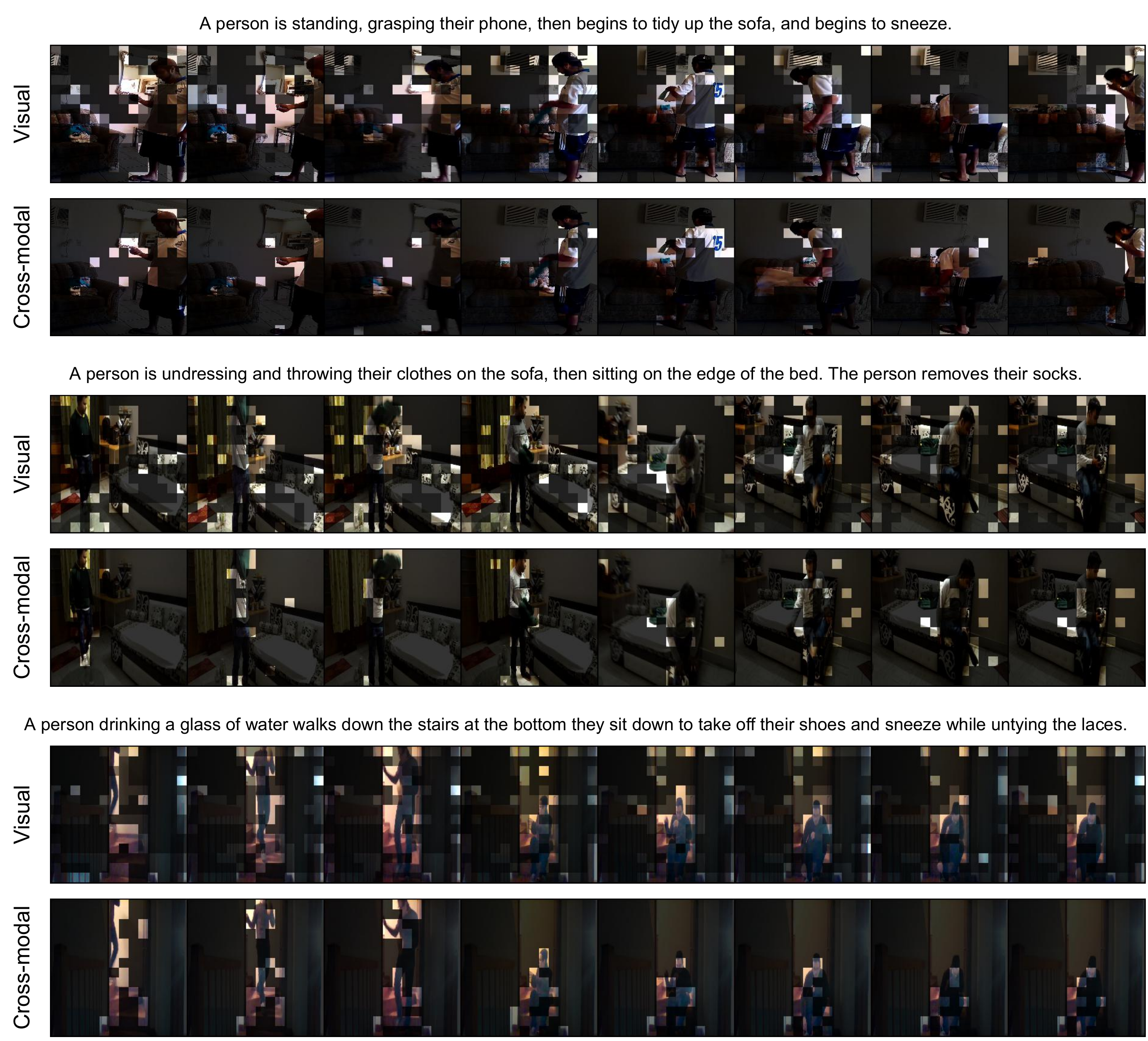}
    \caption{\textbf{Qualitative Results (continued).}}
    \label{fig:supp_qualitative2}
\end{figure*}

%% file: assets/table/supp_token-order.tex
\begin{table}[t]
    \centering

    \resizebox{\linewidth}{!}{
\begin{tabular}{@{}lcccccccc@{}}
\toprule
\multirow{2}{*}{\textbf{Order}} & \multicolumn{4}{c}{\textbf{MSR-VTT}} & \multicolumn{4}{c}{\textbf{DiDeMo}} \\
 & R1 & R5 & R10 & \textbf{Mean} & R1 & R5 & R10 & \textbf{Mean} \\ \midrule
Standard & 21.0 & 43.0 & 51.5 & \textbf{38.5} & 29.1 & 53.5 & 63.1 & \textbf{48.6} \\
Morton & 20.6 & 40.6 & 49.4 & \textbf{36.9} & 27.3 & 51.9 & 61.9 & \textbf{47.1} \\
Hilbert & 20.3 & 40.9 & 49.6 & \textbf{36.9} & 27.9 & 52.5 & 62.5 & \textbf{47.7} \\ \bottomrule
\end{tabular}
    }
    
    \caption{\textbf{Ablation on Token Ordering.} We compare the standard \ours\ trained with flattened video tokens and reordering using space-filling curves.}
    \label{tab:token-order-ablation}
\end{table}

%% file: assets/table/temporal-probe.tex
\begin{table}[t]
    \centering

    \resizebox{\linewidth}{!}{
\begin{tabular}{@{}llcccccc@{}}
\toprule
 &  & \multicolumn{3}{c}{\textbf{SSv2-Label}} & \multicolumn{3}{c}{\textbf{ActivityNet-QA}} \\
\multirow{-2}{*}{\textbf{Model}} & \multirow{-2}{*}{\textbf{Sp.}} & N & S & $\Delta$ & N & S & $\Delta$ \\ \midrule
Singularity~\cite{lei2022revealing} & --- & 66.6 & 66.3 & \textbf{0.3} & 41.8 & 41.8 & \textbf{0.0} \\
 & D & 66.1 & 64.9 & \textbf{1.2} & 42.5 & 42.3 & \textbf{0.2} \\
\multirow{-2}{*}{\ours} & \cellcolor[HTML]{FFF2CC}H & \cellcolor[HTML]{FFF2CC}69.3 & \cellcolor[HTML]{FFF2CC}65.8 & \cellcolor[HTML]{FFF2CC}\textbf{3.5} & \cellcolor[HTML]{FFF2CC}43.2 & \cellcolor[HTML]{FFF2CC}42.3 & \cellcolor[HTML]{FFF2CC}\textbf{0.9} \\ \bottomrule
\end{tabular}
    }
    
    \caption{\textbf{Temporal Probing.} Video-text transformers are evaluated using \underline{N}ormal and \underline{S}huffled frame order.}
    \label{tab:temporal-probe}
\end{table}

%% file: supp/3_discussions.tex
\section{Limitations \& Future Work}
\label{supp:discussions}

While \ours\ shows great potential towards building long-term video-text models, we recognize that learning temporal relationships from videos would not be possible without high-quality pre-training data. We find that WebVid-2M exists a strong tendency towards spatial appearances: Many videos consist of only simple motions (running, talking etc.), and captions are often highly correlated to the static background. Given this, we suspect that further increasing the clip length beyond 16 frames per video is unlikely to make a significant difference in modeling performance. Building on top of the sparse video-text architecture in this work, future studies can focus on pre-training on video-language datasets and tasks that require aggregating information over a longer period of time span, e.g. narrated egocentric videos over long episodes~\cite{grauman2022ego4d}, where \ours\ may provide larger gains over frame-based approaches and dense spatiotemporal transformers. 